\documentclass[sigconf,nonacm]{acmart}
\setcopyright{none}
\acmDOI{}
\acmISBN{}
\AtBeginDocument{%
  }

\acmConference[EvalTrustAgenticAI@KDD'26]
{KDD Workshop on Evaluation and Trustworthiness of Agentic AI}
{August 2026}
{Jeju, Korea}

\usepackage{amsmath}
\usepackage{mathtools}
\usepackage{amssymb}
\usepackage{mathrsfs}
\usepackage{bm}

\newcommand{\Coins}{\mathscr{C}}
\newcommand{\Agents}{\mathscr{A}}
\newcommand{\Cases}{\mathscr{D}}
\newcommand{\Look}{\mathscr{L}}
\newcommand{\Horizon}{\mathscr{H}}
\newcommand{\Labels}{\mathscr{Y}}
\newcommand{\Features}{\bm{x}}
\newcommand{\depeg}{\delta}
\newcommand{\risk}{\varpi}
\newcommand{\conf}{\kappa}

\usepackage{booktabs}
\usepackage{tabularx}
\usepackage{array}
\usepackage{makecell}
\usepackage{multirow}
\usepackage[table]{xcolor}
\usepackage{adjustbox}
\usepackage{threeparttable}

\usepackage{pdflscape}
\usepackage{longtable}
\usepackage{ragged2e}

\usepackage{enumitem}

\usepackage{pgfplots}
\usepgfplotslibrary{groupplots}
\pgfplotsset{compat=1.18}

\usepackage{float}

\usepackage{graphicx}

\usepackage{tikz}
\usetikzlibrary{
  arrows.meta,
  positioning,
  fit,
  backgrounds,
  calc,
  shadows.blur
}

\usepackage{fontawesome5}

\usepackage{hyperref}

\usepackage{xcolor}
\usepackage{listings}

\lstdefinelanguage{XML}{
  morestring=[b]",
  morecomment=[s]{<!--}{-->},
  morekeywords={mxfile,diagram,mxGraphModel,root,mxCell,mxGeometry},
  sensitive=true
}

\definecolor{HeaderGray}{HTML}{F4F4F4}
\definecolor{RowGray}{HTML}{FAFAFA}
\definecolor{LinkPurple}{HTML}{6A1B6A}
\definecolor{OpenClawBlue}{HTML}{1E88E5}
\definecolor{NanoBotOrange}{HTML}{FB8C00}
\definecolor{NanoClawGreen}{HTML}{2ECC71}
\definecolor{HermesTeal}{HTML}{26C6DA}
\definecolor{EvoPurple}{HTML}{8E24AA}
\definecolor{LangGraphGray}{HTML}{546E7A}

\hypersetup{
    colorlinks=true,
    linkcolor=LinkPurple,
    citecolor=LinkPurple,
    urlcolor=LinkPurple
}

\newcommand{\sourcetag}[2]{%
    \href{#1}{\textcolor{LinkPurple}{\underline{#2}}}%
}

\newcommand{\dimicon}{\textcolor{black}{\faTable}}
\newcommand{\openclawicon}{\textcolor{OpenClawBlue}{\faUserAstronaut}}
\newcommand{\nanoboticon}{\textcolor{NanoBotOrange}{\faBolt}}
\newcommand{\nanoclawicon}{\textcolor{NanoClawGreen}{\faIcon{shield-alt}}}
\newcommand{\hermesicon}{\textcolor{HermesTeal}{\faDove}}
\newcommand{\evoicon}{\textcolor{EvoPurple}{\faProjectDiagram}}
\newcommand{\langgraphicon}{\textcolor{LangGraphGray}{\faSitemap}}

\newcolumntype{Y}{>{\RaggedRight\arraybackslash}X}

\makeatletter
\renewcommand{\@fnsymbol}[1]{%
  \ensuremath{%
    \ifcase#1\or \dagger\or \mathsection\or \ddagger\or *\or \mathparagraph\or \|\or **\or \dagger\dagger\or \ddagger\ddagger \else\@ctrerr\fi%
  }%
}
\makeatother

\begin{document}

\title{\textsc{StableEval Arena}: A Cost-Aware Agentic Benchmark for Stablecoin Price Stability Prediction}

\newcommand{\sharedaffiliation}{%
  \affiliation{%
    \institution{Duke Kunshan University}
    \city{Suzhou}
    \country{China}
  }%
}

\author{Sean Wan}
\sharedaffiliation
\authornote{\textbf{Acknowledgments}: Sean Wan is grateful for the support from the Summer Research Scholar Program at Duke Kunshan University, supervised by Prof. Luyao Zhang. }

\author{Dongping Liu}
\affiliation{%
  \institution{Tenorshare}
  \city{Hong Kong}
  \country{China}
}

\authornote{Dongping Liu contributed to this research in his personal capacity and as an independent scholarly endeavor. Tenorshare did not provide funding, resources, institutional support, or endorsement for this work. }

\author{Luyao Zhang}
\sharedaffiliation
\authornotemark[1]
\authornote{The Corresponding author: Email: lz183@duke.edu, Digital Innovation Research Center and Social Science Division, Duke Kunshan University, Address: Duke Avenue No.8, Kunshan, Suzhou, Jiangsu, China, 215316. }

\renewcommand{\shortauthors}{Wan et al.}

\begin{abstract}
We introduce \textsc{StableEval Arena}, a cost-aware benchmark framework for evaluating agentic AI systems on stablecoin peg-risk prediction. 
\textsc{StableEval Arena} evaluates LLM-backed agentic systems on diagnosing peg stress and forecasting deviations from the one-dollar peg over a hidden seven-day horizon, using leakage-safe historical replay with exchange price-volume data and market-context features.
We report two complementary experiment blocks: a 120-case stress-enriched validation block and a 507-case natural-distribution full-arena evaluation block. 
Across six LLM-backed agent configurations and baselines, \textsc{StableEval Arena} measures prediction quality, calibrated-label behavior, structured-output reliability, latency, token consumption, and estimated inference cost. 
Rather than ranking agents by accuracy alone, the framework treats trustworthiness as a joint property of forecast quality, operational reliability, and computational cost. 
The results show a gap between protocol-following reliability and financial-risk reliability: agents reliably produce valid structured outputs at modest measured cost, but still miss most rare severe-stress and sustained-depeg cases.
To support auditing and replication, we release the benchmark dataset on Hugging Face and the source code on GitHub.
\end{abstract}






\begin{teaserfigure}
    \centering
    \makebox[\textwidth][c]{%
        \includegraphics[
            width=1.12\textwidth,
            keepaspectratio
        ]{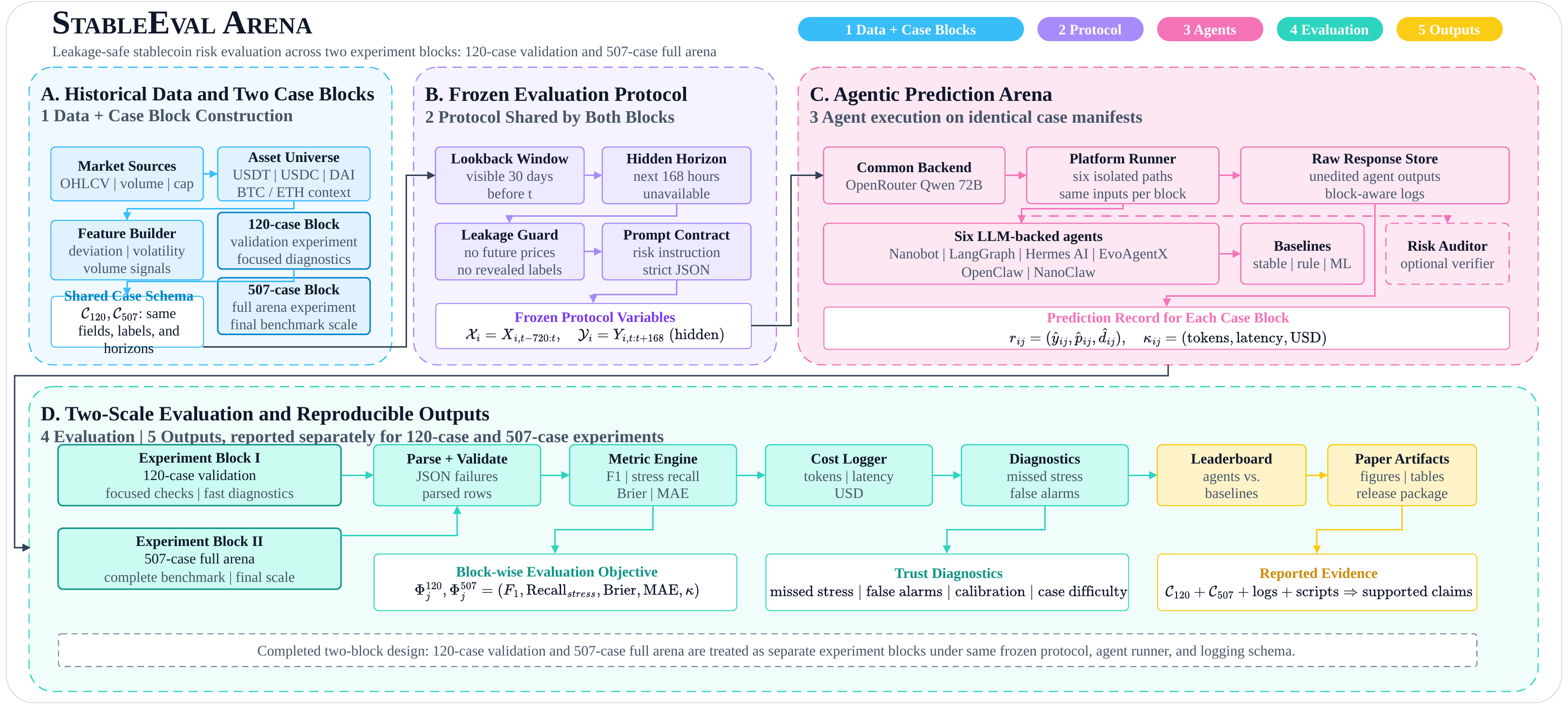}
    }
    \vspace{-2.4em}
    \caption{Overview of \textsc{StableEval Arena}.}
    \Description{Overview diagram of the StableEval Arena workflow, including data construction, frozen protocol design, agent execution, evaluation, cost analysis, diagnostics, and reproducible artifacts.}
    \label{fig:stableeval-arena-overview}
\end{teaserfigure}

\maketitle

\vspace{-1em}
\section{Introduction}
\label{sec:introduction}

Stablecoins are core digital finance infrastructure for trading, lending, payments, collateral, liquidity, and settlement.
Their utility rests on a fragile promise: a dollar-pegged stablecoin should remain near one U.S. dollar. 
When the promise weakens, applications may face pricing errors, liquidity stress, payment failure, and settlement uncertainty. 
Recent stablecoin research therefore calls for transparent metrics, stakeholder-oriented evaluation, and reproducible benchmarks~\cite{zhang2025sokstablecoins}. 
Yet stablecoin risk assessment differs from ordinary cryptocurrency forecasting: the goal is not to predict the full future price path, but to judge whether a stablecoin will remain stable, enter a warning state, or experience peg stress over a fixed future window. 
Such judgments require bounded evidence about peg deviation, volatility, volume, market context, and stablecoin mechanisms.
Cryptocurrency-forecasting work highlights multimodal financial time-series inputs, including market metrics and sentiment signals~\cite{fu2024dam}, but stablecoin evaluation must separate normal peg noise from rare, high-consequence stress events.

This evaluation problem becomes more demanding as agentic AI systems enter financial decision-support workflows, including autonomous prediction-market trading~\cite{zhang2026predictionarena}.
These systems increasingly support reasoning, tool use, workflow execution, and decision support, yet trustworthy evaluation requires more than final-answer accuracy.
Benchmarks must also measure protocol compliance, output validity, runtime cost, reliability, and failure modes. 
Recent agent and prediction-market benchmarks motivate standardized execution, fixed budgets, process instrumentation, and efficiency metrics for autonomous systems~\cite{chen2026agent2rlbench,zhang2026predictionarena}. 
Related forecasting and tabular benchmarks emphasize unified evaluation protocols, public leaderboards, reproducible code, and maintained benchmark infrastructure~\cite{li2025tsfmbench,erickson2025tabarena}.
However, existing benchmarks do not directly test whether agentic AI systems can serve as stablecoin risk-assessment agents under cost and reliability constraints. 

We address this gap with \textsc{StableEval Arena}, a historical-replay benchmark for evaluating agentic stablecoin risk assessment.
Each case provides a frozen 30-day lookback window and asks for a structured next-7-day risk assessment: stable/watch/stress label, depeg probability, maximum peg-deviation estimate, direction, confidence, and rationale. 
We compare agentic systems and non-agent baselines under one protocol, measuring prediction quality, stress recall, missed-depeg risk, calibration, valid-output rate, latency, token usage, and inference cost. 
The benchmark has two settings: a 120-case stress-enriched validation block for stress diagnosis and a 507-case natural-distribution scaling block for full-pool evaluation.

We organize the study around three research questions:
\begin{itemize}[leftmargin=*, topsep=2pt, itemsep=1pt, parsep=0pt, partopsep=0pt]
    \item \textbf{RQ1 (comparative performance).} Under a shared historical-replay protocol, do agentic AI frameworks improve stablecoin risk forecasts over simple baselines, or reproduce same errors?
    \item \textbf{RQ2 (reliability under rare stress).} When depeg stress is uncommon, how do systems trade off missed stress events, false alarms, calibration, structured-output validity, latency, token usage, and inference cost?
    \item \textbf{RQ3 (benchmark sensitivity).} What does comparing a stress-enriched validation block with a natural-distribution scaling block reveal that a single aggregate leaderboard would obscure?
\end{itemize}

The results show a gap between procedural reliability and financial-risk detection. The benchmark is diagnostic rather than agent-promotional: agents produce valid structured outputs under shared protocol, but do not consistently outperform simple baselines. 
In stress-enriched validation block, calibrated agents improve aggregate Macro-F1 relative to simple baselines; in the 507-case natural-distribution block, simple baselines remain strong and raw accuracy is shaped by stable-dominant class balance. 
Across both settings, rare-stress and sustained-depeg detection remain weak. To support replication, we release the dataset on Hugging Face\footnote{\textit{\url{https://huggingface.co/datasets/SeanWan05/stableeval-arena}}} and the source code on GitHub\footnote{\textit{\url{https://github.com/StableTradeAtlas/StableEval-Arena}}} as open-access resources.

\enlargethispage{2\baselineskip}
\section{Background and Related Work}
\label{sec:background&relatedwork}

\definecolor{stableNavy}{HTML}{0F172A}
\definecolor{stableDark}{HTML}{334155}
\definecolor{stableGray}{HTML}{64748B}
\definecolor{stableBg}{HTML}{F8FAFC}

\definecolor{stableData}{HTML}{38BDF8}
\definecolor{stableDataCard}{HTML}{E0F2FE}

\definecolor{stableProtocol}{HTML}{A78BFA}
\definecolor{stableProtocolCard}{HTML}{F3E8FF}

\definecolor{stableEval}{HTML}{2DD4BF}
\definecolor{stableEvalCard}{HTML}{CCFBF1}

\definecolor{stableAmber}{HTML}{FACC15}
\definecolor{stableAmberCard}{HTML}{FEF9C3}

\definecolor{stableRisk}{HTML}{F472B6}
\definecolor{stableRiskCard}{HTML}{FDF2F8}

\begin{table*}[t]
\centering
\caption{\textbf{Agentic Systems Selected for StableEval Arena Evaluation}}
\label{tab:agentic_ai_comparison_refined}

\vspace{-0.8em}
{\small
This table summarizes the six agentic systems included as StableEval Arena comparators, focusing on implementation role, deployment interface, memory/persistence, orchestration, and control features relevant to benchmark execution.
}

\vspace{0.7em}

\small
\setlength{\tabcolsep}{4.5pt}
\renewcommand{\arraystretch}{1.18}
\arrayrulecolor{stableGray!65}

\begin{threeparttable}
\begin{adjustbox}{max width=\textwidth}
\begin{tabularx}{\textwidth}{
    >{\raggedright\arraybackslash\bfseries}p{1.85cm}
    Y Y Y Y Y Y
}

\arrayrulecolor{stableData}
\toprule[1.25pt]
\rowcolor{white}
\makecell[c]{\rule{0pt}{3.2ex}\textcolor{stableData}{\dimicon}\quad \textbf{Dimension}}
&
\makecell[c]{\rule{0pt}{3.2ex}\textcolor{stableProtocol}{\nanoboticon}\quad \textbf{Nanobot}}
&
\makecell[c]{\rule{0pt}{3.2ex}\textcolor{stableProtocol}{\langgraphicon}\quad \textbf{LangGraph}}
&
\makecell[c]{\rule{0pt}{3.2ex}\textcolor{stableProtocol}{\hermesicon}\quad \textbf{Hermes AI}}
&
\makecell[c]{\rule{0pt}{3.2ex}\textcolor{stableProtocol}{\evoicon}\quad \textbf{EvoAgentX}}
&
\makecell[c]{\rule{0pt}{3.2ex}\textcolor{stableProtocol}{\openclawicon}\quad \textbf{OpenClaw}}
&
\makecell[c]{\rule{0pt}{3.2ex}\textcolor{stableProtocol}{\nanoclawicon}\quad \textbf{NanoClaw}}
\\[0.4em]

\arrayrulecolor{stableData}
\midrule[0.95pt]

\rowcolor{stableDataCard}
System Type
&
Ultra-lightweight personal AI agent
&
Graph-based agent workflow framework
&
Self-improving-oriented personal AI agent
&
Evolving multi-agent workflow framework
&
General-purpose AI assistant
&
OpenClaw-family alternative
\\

\arrayrulecolor{stableData!70}
\midrule

\rowcolor{stableProtocolCard}
Access / License
&
Open-source; MIT 
\sourcetag{https://github.com/HKUDS/nanobot}{[NB-GH 2026]}
&
OS LangChain ecosystem
\sourcetag{https://github.com/langchain-ai/langgraph}{[LG-GH 2026]};
\sourcetag{https://docs.langchain.com/oss/python/langgraph/overview}{[LG-Docs 2026]}
&
Open-source; MIT 
\sourcetag{https://github.com/NousResearch/hermes-agent}{[HA-GH 2026]}
&
Open-source; MIT; paper available 
\sourcetag{https://github.com/EvoAgentX/EvoAgentX}{[EAX-GH 2026]};
\sourcetag{https://arxiv.org/abs/2507.03616}{[Wang et al. 2025]}
&
Open-source; MIT 
\sourcetag{https://github.com/openclaw/openclaw}{[OC-GH 2026]}
&
Open-source; MIT 
\sourcetag{https://github.com/qwibitai/nanoclaw}{[NC-GH 2026]}
\\

\arrayrulecolor{stableProtocol!70}
\midrule

\rowcolor{stableEvalCard}
Interface / Deployment
&
CLI, gateway, chat apps, Docker/Linux
&
Python framework; graph workflows; deploy support
&
CLI plus messaging gateway; multi-backend terminals
&
Python framework, notebooks, MCP/tool workflows
&
Multi-channel assistant; desktop/mobile; gateway
&
Messaging apps via skills; container-first runtime
\\

\arrayrulecolor{stableEval!70}
\midrule

\rowcolor{stableDataCard}
Memory and Persistence
&
Long-term memory, Dream memory, heartbeat schedule
&
Short-/long-term memory; durable execution
&
Persistent memory, user modeling, self-improving skills
&
Short-/long-term memory modules
&
Sessions, skills, cron, usage commands
&
Group memory and scheduled jobs
\\

\arrayrulecolor{stableData!70}
\midrule

\rowcolor{stableProtocolCard}
Tools / Orchestration
&
Small agent loop; tools, skills, MCP
&
Graph control, tools, HITL, persistence
&
Tools, skills, cron, isolated subagents, MCP
&
Workflow plus TextGrad, AFlow, MIPRO
&
Skills, model failover, multi-agent routing
&
Skills, web access, Anthropic Agents SDK
\\

\arrayrulecolor{stableProtocol!70}
\midrule

\rowcolor{stableRiskCard}
Safety / Control
&
Hardened shell/tool behavior; Docker deployment
&
Human-in-the-loop control and durable state
&
Command approval, DM pairing, container isolation
&
HITL checkpoints and tool control
&
Optional Docker sandbox for non-main sessions
&
Container isolation and credential separation
\\

\arrayrulecolor{stableRisk}
\bottomrule[1.25pt]

\end{tabularx}
\end{adjustbox}

\vspace{0.35em}
\begin{tablenotes}[flushleft]
\scriptsize
\setlength{\itemsep}{0.2em}

\item
\colorbox{stableBg}{%
\parbox{\dimexpr\textwidth-2\fboxsep\relax}{
\textbf{Note.}
Rows summarize system type, access, interface, memory, orchestration, and safety/control dimensions used to characterize comparable agentic AI systems. Categories reflect how recent surveys describe LLM-based agents as systems combining planning, memory, tool use, feedback, and evaluation/control.

\textbf{Selection.}
Nanobot, LangGraph, Hermes AI, and EvoAgentX represent native/framework-level agent designs; OpenClaw and NanoClaw are treated as adapter-based OpenClaw-family comparators.

\textbf{Source tags.}
NB-GH: Nanobot GitHub;
LG-GH/LG-Docs: LangGraph GitHub/docs;
HA-GH: Hermes Agent GitHub;
EAX-GH: EvoAgentX GitHub;
Wang et al. 2025: \textit{EvoAgentX};
OC-GH: OpenClaw GitHub;
NC-GH: NanoClaw GitHub.
}
}

\end{tablenotes}

\end{threeparttable}
\arrayrulecolor{black}
\vspace{-1em}
\end{table*}

\subsection{Stablecoin Risk Monitoring}

Stablecoin risk monitoring asks whether a token's peg is supported by mechanisms that remain reliable under stress, not whether its recent price has stayed close to par. 
Prior work shows that stablecoin stability depends on market design and arbitrage access~\cite{lyons2023stablecoins}, reserve or collateral quality, redemption mechanisms, liquidity, user confidence, and regulatory design~\cite{arner2020stablecoins,gorton2023taming,klagesmundt2020stablecoins}.
These factors matter because a token can appear stable in ordinary conditions while still being exposed to liquidity shocks, redemption frictions, or confidence-driven runs. 
Monitoring therefore requires distinguishing routine peg noise from warning or stress conditions that may affect payments, liquidation thresholds, collateral valuation, and settlement reliability. 
Recent work further argues that stablecoins should be evaluated as programmable financial infrastructure shaped by stakeholder needs, governance, regulation, and transparent benchmarking~\cite{zhang2025sokstablecoins}. 
This motivates \textsc{StableEval Arena}'s framing of monitoring as an evidence-bounded risk-evaluation task: given limited information, a system should identify salient risk signals, classify the severity of peg conditions, and produce reproducible assessments grounded in the available evidence.

\vspace{-1.2em}
\subsection{Agentic Financial Prediction and Evaluation}

Agentic financial prediction extends beyond conventional time-series forecasting because the system is often asked to synthesize bounded evidence into a structured judgment that can support downstream decisions. 
Time-series benchmarks emphasize fixed protocols, standardized dataset handling, and comparable evaluation settings~\cite{li2025tsfmbench}, while cryptocurrency forecasting research shows that multimodal evidence can add value beyond price-only modeling~\cite{fu2024dam}. 
In financial-risk settings, however, prediction quality alone is insufficient: the reliability of the decision process also depends on whether the system follows instructions, respects the evaluation protocol, produces valid structured outputs, and behaves consistently under resource constraints. 
This concern connects financial prediction to agent-evaluation work, where benchmarks such as AgentBench highlight reasoning, decision-making, and instruction following as persistent challenges~\cite{liu2023agentbench}, and Agent2 RL-Bench emphasizes fixed-budget evaluation, runtime recording, failure diagnosis, and post-hoc analysis~\cite{chen2026agent2rlbench}. 
These strands motivate evaluating agentic financial systems not only by predictive performance, but also by protocol compliance, structured-output reliability, failure behavior, latency, and resource use.

\vspace{-1.2em}
\subsection{Cost-Aware and Trustworthy Benchmarking}

Cost-aware and trustworthy benchmarking asks whether a system is reliable under the operational conditions in which its predictions would be used, not only whether it achieves high average accuracy. 
This distinction is important in financial settings, where a model can appear strong on aggregate metrics while missing rare but consequential stress events, producing malformed outputs, incurring excessive latency, or failing to reproduce results under a fixed protocol. 
Recent work on prediction-market agents reflects this broader view: Prediction Arena evaluates autonomous AI models using returns and win rates alongside token usage, cycle time, settlement behavior, exit patterns, and cross-platform differences~\cite{zhang2026predictionarena}. 
For stablecoin risk monitoring, this perspective suggests that evaluation should account for both decision quality and the cost and reliability of obtaining that decision, especially when agents are queried repeatedly across assets and time windows.

Trustworthiness also requires diagnostics sensitive to imbalanced risk settings. 
In stable-dominant domains, high accuracy may reflect majority-class behavior while rare depeg or stress cases remain undetected.
Evaluation therefore benefits from reporting stress recall, missed-depeg rates, false alarms, calibration, maximum-deviation error, inference cost, latency, and failure modes alongside aggregate performance.
Benchmark-infrastructure work emphasizes reproducible, transparent documentation: TabArena highlights curated datasets, public code, leaderboard-style reporting, and explicit limitations~\cite{erickson2025tabarena}, while Datasheets for Datasets motivates reporting dataset provenance, composition, intended uses, limitations, maintenance, and risks or harms~\cite{gebru2021datasheets}.
These principles provide the background for \textsc{StableEval Arena} benchmark, where financial-agent evaluation should report prediction quality, cost, reproducibility, and stress-specific diagnostics.

\vspace{-1.5em}
\section{\textsc{StableEval Arena}}
\label{sec:stableeval}

\subsection{Overview}

\textsc{StableEval Arena} is a historical-replay benchmark for trustworthy, cost-aware evaluation of agentic AI systems in stablecoin risk assessment. 
Specifically, each system receives evidence available at an observation time, and must produce a structured next-window risk assessment: whether the target stablecoin remains stable, enters a warning state, or experiences peg stress, together with its estimated depeg probability, maximum dollar-peg deviation, deviation direction, confidence, and rationale.
The benchmark is designed to support fair comparison, financial relevance, and auditability: all systems receive same agent-facing inputs and output schema; labels correspond to short-horizon stablecoin risk; each run records outputs, cost, latency, failures, and reproducibility metadata.

\begin{figure*}[t]
    \centering
\definecolor{stableNavy}{HTML}{0F172A}
\definecolor{stableBg}{HTML}{F8FAFC}

\definecolor{stableData}{HTML}{38BDF8}
\definecolor{stableDataPanel}{HTML}{F0F9FF}
\definecolor{stableDataCard}{HTML}{E0F2FE}

\definecolor{stableProtocol}{HTML}{A78BFA}
\definecolor{stableProtocolPanel}{HTML}{FAF5FF}
\definecolor{stableProtocolCard}{HTML}{F3E8FF}

\definecolor{stableEval}{HTML}{2DD4BF}
\definecolor{stableEvalPanel}{HTML}{F0FDFA}
\definecolor{stableEvalCard}{HTML}{CCFBF1}

\definecolor{stableAmber}{HTML}{FACC15}
\definecolor{stableAmberCard}{HTML}{FEF9C3}

\definecolor{stableRisk}{HTML}{F472B6}
\definecolor{stableRiskCard}{HTML}{FDF2F8}

\definecolor{stableBlock}{HTML}{D97706}
\definecolor{stableBlockCard}{HTML}{FEF3C7}

\definecolor{stableGray}{HTML}{64748B}

\resizebox{\textwidth}{!}{%
\begin{tikzpicture}[
    font=\rmfamily,
    >=Latex,
    panel/.style 2 args={
        rounded corners=8pt,
        draw=#1,
        fill=#2,
        line width=1.15pt,
        minimum width=4.35cm,
        minimum height=5.55cm
    },
    paneltitle/.style 2 args={
        rounded corners=6pt,
        draw=#1,
        fill=#2,
        line width=0.9pt,
        align=center,
        text=stableNavy,
        font=\rmfamily\bfseries\normalsize,
        minimum width=3.85cm,
        minimum height=0.74cm
    },
    item/.style 2 args={
        rounded corners=5pt,
        draw=#1,
        fill=#2,
        line width=0.75pt,
        align=center,
        text=stableNavy,
        font=\rmfamily\scriptsize,
        text width=1.75cm,
        minimum height=0.76cm,
        inner sep=1.5pt
    },
    mainarrow/.style={
        -{Latex[length=2.5mm]},
        line width=1.05pt,
        draw=stableNavy!70
    },
    arrowlabel/.style={
        fill=stableBg,
        inner sep=1pt,
        font=\rmfamily\bfseries\tiny,
        text=stableNavy
    }
]

\node[panel={stableData}{stableDataPanel}]         (inputPanel)  at (0,0) {};
\node[panel={stableProtocol}{stableProtocolPanel}] (methodPanel) at (5.15,0) {};
\node[panel={stableEval}{stableEvalPanel}]         (outputPanel) at (10.30,0) {};

\node[paneltitle={stableData}{stableDataCard}] (inputTitle) at (0,2.17)
{\faDatabase\quad Inputs};

\node[paneltitle={stableProtocol}{stableProtocolCard}] (methodTitle) at (5.15,2.17)
{\faCogs\quad Methodology};

\node[paneltitle={stableEval}{stableEvalCard}] (outputTitle) at (10.30,2.17)
{\faChartBar\quad Outputs};

\node[item={stableBlock}{stableBlockCard}] (i_cases120) at (-0.98,1.25)
{\faBullseye\\120 Enriched\\Cases};

\node[item={stableBlock}{stableBlockCard}] (i_cases507) at (0.98,1.25)
{\faLayerGroup\\507 Natural\\Cases};

\node[item={stableData}{stableDataCard}] (i_assets) at (-0.98,0.42)
{\faCoins\\Stablecoins\\USDT/USDC/DAI};

\node[item={stableData}{stableDataCard}] (i_window) at (0.98,0.42)
{\faClock\\30-Day\\Lookback};

\node[item={stableData}{stableDataCard}] (i_features) at (-0.98,-0.41)
{\faChartLine\\Risk\\Features};

\node[item={stableData}{stableDataCard}] (i_context) at (0.98,-0.41)
{\faGlobe\\BTC/ETH\\Context};

\node[item={stableData}{stableDataCard}] (i_agents) at (-0.98,-1.24)
{\faRobot\\Agentic\\Systems};

\node[item={stableData}{stableDataCard}] (i_baselines) at (0.98,-1.24)
{\faRulerCombined\\Baseline\\Suite};

\node[item={stableBlock}{stableBlockCard}] (m_120) at (4.17,1.25)
{\faBullseye\\120 Enriched\\Block};

\node[item={stableBlock}{stableBlockCard}] (m_507) at (6.13,1.25)
{\faLayerGroup\\507 Natural\\Block};

\node[item={stableProtocol}{stableProtocolCard}] (m_packet) at (4.17,0.42)
{\faClipboardList\\Prompt\\Packet};

\node[item={stableProtocol}{stableProtocolCard}] (m_schema) at (6.13,0.42)
{\faCode\\Strict\\JSON};

\node[item={stableProtocol}{stableProtocolCard}] (m_protocol) at (4.17,-0.41)
{\faBalanceScale\\Unified\\Protocol};

\node[item={stableProtocol}{stableProtocolCard}] (m_backend) at (6.13,-0.41)
{\faServer\\Qwen\\Backend};

\node[item={stableProtocol}{stableProtocolCard}] (m_calib) at (4.17,-1.24)
{\faSlidersH\\Label\\Calibration};

\node[item={stableProtocol}{stableProtocolCard}] (m_metrics) at (6.13,-1.24)
{\faClipboardCheck\\Metric\\Engine};

\node[item={stableBlock}{stableBlockCard}] (o_120) at (9.32,1.25)
{\faBullseye\\120 Enriched\\Results};

\node[item={stableBlock}{stableBlockCard}] (o_507) at (11.28,1.25)
{\faLayerGroup\\507 Natural\\Results};

\node[item={stableEval}{stableEvalCard}] (o_leader) at (9.32,0.42)
{\faTrophy\\Grouped\\Leaderboard};

\node[item={stableRisk}{stableRiskCard}] (o_stress) at (11.28,0.42)
{\faExclamationTriangle\\Stress\\Audit};

\node[item={stableEval}{stableEvalCard}] (o_align) at (9.32,-0.41)
{\faBalanceScale\\120--507\\Alignment};

\node[item={stableEval}{stableEvalCard}] (o_cost) at (11.28,-0.41)
{\faDollarSign\\Cost\\Latency};

\node[item={stableEval}{stableEvalCard}] (o_logs) at (9.32,-1.24)
{\faFingerprint\\Logs\\Hashes};

\node[item={stableEval}{stableEvalCard}] (o_claims) at (11.28,-1.24)
{\faIcon{file-alt}\\Paper\\Claims};

\draw[mainarrow]
    (inputPanel.east) -- node[arrowlabel, above=1pt] {frozen packets}
    (methodPanel.west);

\draw[mainarrow]
    (methodPanel.east) -- node[arrowlabel, above=1pt] {validated scores}
    (outputPanel.west);

\node[
    rounded corners=5pt,
    draw=stableRisk,
    fill=stableRiskCard,
    line width=0.8pt,
    align=center,
    text=stableNavy,
    font=\rmfamily\scriptsize\bfseries,
    text width=13.35cm,
    minimum width=13.75cm,
    minimum height=0.58cm,
    inner sep=2pt
] (trust) at (5.15,-2.28)
{\faUserShield\quad Trustworthiness lens: rare-stress recall, missed-depeg risk, false alarms, calibration, cost, latency, failure rate, and reproducibility.};

\end{tikzpicture}%
}
    \vspace{-1.8em}
    \caption{Experiment design of \textsc{StableEval Arena}: inputs, evaluation protocol, and reported outputs.}
    \label{fig:stableeval-experiment-design}
    \vspace{-1.5em}
\end{figure*}
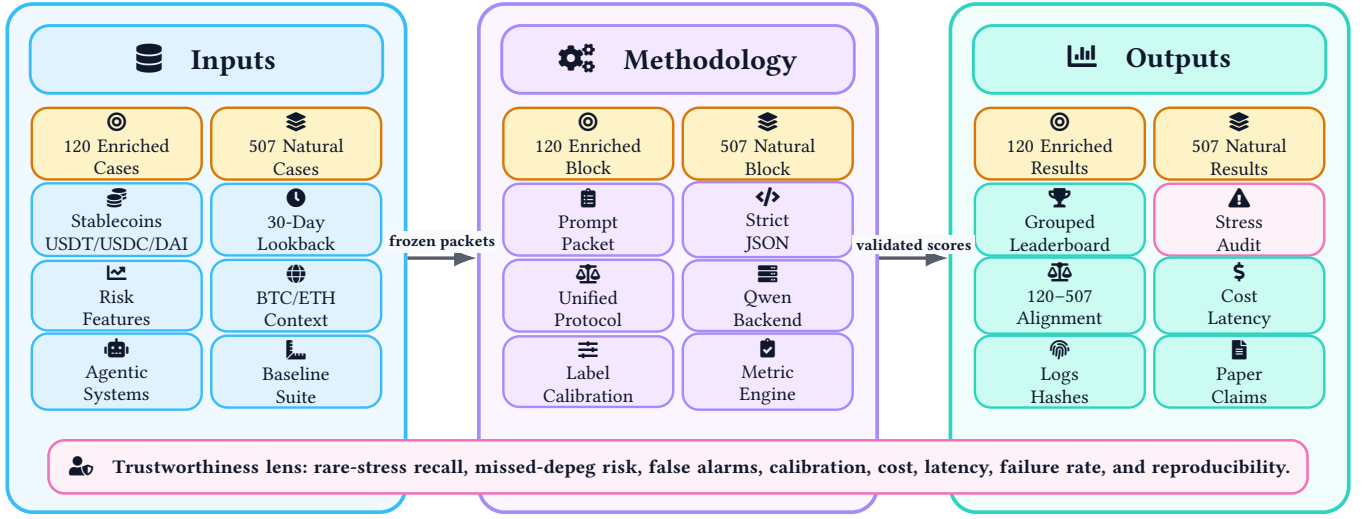

The completed \textsc{StableEval Arena} benchmark uses two complementary evaluation settings. 
We refer to the full natural-distribution setting as \textsc{StableEval-507-Natural}: a 507-case historical-replay pool used for robustness and full-pool evaluation. 
We refer to the stress-enriched diagnostic setting as \textsc{StableEval-120-Enriched}: a 120-case subset that concentrates rare watch and stress outcomes for sharper stress analysis. 
\textsc{StableEval-507-Natural} contains 441 stable, 45 watch, and 21 stress cases, while \textsc{StableEval-120-Enriched} contains 54 stable, 45 watch, and 21 stress cases. 
Thus, the natural distribution benchmark is centered on \textsc{StableEval-507-Natural}, and stress diagnosis is centered on \textsc{StableEval-120-Enriched}. 
As summarized in Figure~\ref{fig:stableeval-experiment-design}, the two settings are complementary rather than interchangeable.
\textsc{StableEval-507-Natural} tests behavior under natural class balance, whereas \textsc{StableEval-120-Enriched} tests whether systems respond to sparse but consequential stress cases.

\subsection{Task Formulation}

Each benchmark case in \textsc{StableEval Arena} is defined by a target stablecoin, an observation time $t$, a frozen 30-day lookback window with hourly data, and a hidden 7-day prediction horizon. 
The target stablecoins are USDT, USDC, and DAI. 
The lookback window contains 720 hours of evidence, and the hidden horizon contains the next 168 hours. 
Notice that the agent is not asked to reproduce the full future hourly price path. 
Instead, it predicts a structured next-window risk assessment: stable/watch/stress class, depeg probability, maximum peg-deviation estimate, direction, confidence, and rationale.

Formally, let $\Coins=\{\mathrm{USDT},\mathrm{USDC},\mathrm{DAI}\}$ denote the target stablecoin universe and let $\Labels=\{\texttt{stable},\texttt{watch},\texttt{stress}\}$ denote the stability-label space. 
Therefore, a benchmark case $\xi_i\in\Cases$ is defined as
\[
\xi_i=(\chi_i,t_i,\Look_i,\Horizon_i),\qquad \chi_i\in\Coins,
\]
where $\chi_i$ is the target stablecoin, $t_i$ is the observation time, $\Look_i$ is the lookback evidence window, and $\Horizon_i$ is the hidden future horizon. 
The two temporal windows are
\[
\begin{gathered}
\Look_i=\{\Features_{\chi_i,\tau}: \tau=t_i-719,\ldots,t_i\},\\[0.4em]
\Horizon_i=\{p_{\chi_i,\tau}: \tau=t_i+1,\ldots,t_i+168\}.
\end{gathered}
\]
Here $\Features_{\chi,\tau}\in\mathbb{R}^{d}$ contains price, volume, peg-deviation, volatility, liquidity, and market-context features available at hour $\tau$, and $p_{\chi,\tau}$ is the hourly close price.

The required system output is a structured JSON object corresponding to the prediction vector
\[
f_{\theta_a}(\xi_i)\mapsto 
\widehat{\bm{y}}_{a,i}
=
(\widehat{s}_{a,i},\widehat{\risk}_{a,i},\widehat{\Delta}_{a,i},
\widehat{\phi}_{a,i},\widehat{\conf}_{a,i},r_{a,i}),
\]
where $a\in\Agents$ is an evaluated system, $\theta_a$ denotes its platform and backend configuration, $\widehat{s}_{a,i}\in\Labels$ is the predicted stability class, $\widehat{\risk}_{a,i}\in[0,1]$ is the predicted next-window depeg-risk probability, $\widehat{\Delta}_{a,i}\in\mathbb{R}_{\ge 0}$ is the predicted maximum absolute peg deviation in basis points, $\widehat{\phi}_{a,i}\in\{-1,0,+1\}$ is the predicted deviation direction, $\widehat{\conf}_{a,i}\in[0,1]$ is the reported confidence, and $r_{a,i}$ is a short evidence-based rationale. 
Thus, \textsc{StableEval Arena} evaluates agentic decision systems rather than trained time-series forecasters: systems are scored by accuracy, schema validity, calibration, rare-stress behavior, cost, latency, and reliability.

\subsection{Case Library and Construction}

\textsc{StableEval Arena} focuses on USDT, USDC, and DAI, covering centralized issuer-backed and decentralized collateral-backed stablecoins. 
BTC and ETH are included only as context assets to capture broader crypto-market stress. 
Each case is constructed from hourly price-volume data and daily market-structure features. 
The core raw fields include open, high, low, close, base volume, and USD volume. 
From these data, the benchmark derives risk-sensitive features such as latest close price, latest peg deviation, recent maximum absolute deviation, duration above deviation thresholds, deviation volatility, recent high-low range, volume changes, market-cap or supply changes when available, and BTC/ETH market context. 
All agent-facing features are computed only from the 30-day (720-hours) lookback window.

As defined above, \textsc{StableEval-507-Natural} is the full case library, while \textsc{StableEval-120-Enriched} is a stress-enriched diagnostic subset drawn from it. 
The subset retains all watch and stress cases and adds stable controls, so stress behavior can be examined without changing the underlying historical-replay protocol. 
This design lets the benchmark report both natural-distribution results and concentrated stress diagnostics. 
Hidden future labels and horizon statistics are used only for case selection and evaluation; they are never included in the agent-facing prompt packets.

Dollar-peg deviation is measured as absolute distance from the one-dollar reference value. 
For stablecoin $\chi$ at hour $\tau$, we define
\[
\depeg_{\chi,\tau}=10^{4}\lvert p_{\chi,\tau}-1\rvert ,
\]
where $p_{\chi,\tau}$ is the hourly close price. 
For case $\xi_i$, the hidden-horizon maximum deviation and threshold-duration statistic are
\[
\Delta_i=\max_{\tau\in\Horizon_i}\depeg_{\chi_i,\tau},
\qquad
d_i(\vartheta)=\sum_{\tau\in\Horizon_i}\mathbf{1}\{\depeg_{\chi_i,\tau}\ge \vartheta\},
\]
where $\vartheta$ is a pre-declared deviation threshold in basis points. 
We also record a strict sustained-depeg indicator
\[
z_i=\mathbf{1}\left\{\exists \tau:\tau,\tau+1,\tau+2\in\Horizon_i
\ \text{and}\ 
\min_{k\in\{0,1,2\}}\depeg_{\chi_i,\tau+k}\ge 100\right\}.
\]
The final hidden label $\ell_i\in\Labels$ is assigned by the frozen evaluator:
\[
\ell_i=
\begin{cases}
\texttt{stress}, & z_i=1\ \text{or}\ \Delta_i\ge \vartheta_{\mathrm{stress}},\\
\texttt{watch}, & \vartheta_{\mathrm{watch}}\le \Delta_i<\vartheta_{\mathrm{stress}}
\ \text{or}\ d_i(\vartheta_{\mathrm{watch}})\ge h_{\mathrm{watch}},\\
\texttt{stable}, & \text{otherwise},
\end{cases}
\]
where $\vartheta_{\mathrm{watch}}$, $\vartheta_{\mathrm{stress}}$, and $h_{\mathrm{watch}}$ are fixed before evaluation. 
This rule avoids labeling ordinary peg noise as stress while preserving sensitivity to rare but consequential deviations.

\subsection{Agent Systems}

\textsc{StableEval Arena} evaluates the heterogeneous agentic systems summarized in Table~\ref{tab:agentic_ai_comparison_refined} under a shared task interface.
The systems of native/framework-level are Nanobot, LangGraph, Hermes AI, and EvoAgentX. 
These systems form the main agentic AI comparison in the 507-case study. 
OpenClaw and NanoClaw are using StableEval adapter executions.
Since verified local CMDOP execution was not available for OpenClaw and NanoClaw, their results are interpreted as adapter-based -Claw extensions that center on testing protocol compatibility, output validity, and cost logging under the same StableEval interface.

The benchmark also includes non-agent baselines: an always-stable baseline, a historical-persistence baseline, a threshold-rule baseline, and a simple machine-learning baseline where feasible. 
Baselines are included to prevent agent-only comparison and to test whether agentic execution adds value beyond simple rules that exploit the strong base-rate stability of major stablecoins. 
Both \textsc{StableEval-120-Enriched} and \textsc{StableEval-507-Natural} use the same system grouping: non-agent baselines, four native/framework-level agents, and two adapter-based -Claw extensions.
Each reported result includes an execution-mode label, allowing the executions to be interpreted accurately.

\subsection{Evaluation Protocol, Outputs, and Metrics}

All systems are evaluated under a unified protocol. 
They receive the same case packets, use the same prompt contract, follow the same strict JSON schema, and are parsed and scored by the same evaluator. 
The common backend is \texttt{qwen/qwen-2.5-72b-instruct} via OpenRouter with temperature 0. 
Web browsing is disabled, and prompts instruct systems to use only the provided lookback-window evidence. 
The protocol is fixed across the 120-case stress-enriched validation and the 507-case natural-distribution study.

The JSON output schema requires the predicted class $\widehat{s}_{a,i}$, depeg-risk probability $\widehat{\risk}_{a,i}$, maximum deviation estimate $\widehat{\Delta}_{a,i}$, deviation direction $\widehat{\phi}_{a,i}$, confidence $\widehat{\conf}_{a,i}$, and rationale $r_{a,i}$. 
A pre-declared calibration rule maps model-reported probability and predicted deviation into the final stable/watch/stress label. 
Raw and calibrated outputs are retained separately. 
Accuracy is reported, but it is not the primary measure under the stable-dominant 507-case distribution; Macro-F1, balanced accuracy, stress recall, missed-depeg rate, calibration, and deviation error are the primary reported metrics.

Let $N=|\Cases|$ and let $\widehat{s}_{a,i}$ and $\ell_i$ denote the predicted and hidden labels for system $a$ on case $i$. 
For class $y\in\Labels$, precision, recall, and F1 are
\[
P_y=\frac{\mathrm{TP}_y}{\mathrm{TP}_y+\mathrm{FP}_y},\qquad
R_y=\frac{\mathrm{TP}_y}{\mathrm{TP}_y+\mathrm{FN}_y},\qquad
F1_y=\frac{2P_yR_y}{P_y+R_y}.
\]
Macro-F1 and balanced accuracy are then
\[
\mathrm{MacroF1}_a=\frac{1}{|\Labels|}\sum_{y\in\Labels}F1_y,
\qquad
\mathrm{BalAcc}_a=\frac{1}{|\Labels|}\sum_{y\in\Labels}R_y.
\]
For severe-risk evaluation, we define
\[
\begin{gathered}
\mathrm{MDR}_a=
\frac{\sum_i \mathbf{1}\{z_i=1,\widehat{s}_{a,i}\ne \texttt{stress}\}}
{\sum_i \mathbf{1}\{z_i=1\}},\\[0.6em]
\mathrm{FAR}_a=
\frac{\sum_i \mathbf{1}\{z_i=0,\widehat{s}_{a,i}= \texttt{stress}\}}
{\sum_i \mathbf{1}\{z_i=0\}},
\end{gathered}
\]
where $\mathrm{MDR}_a$ is the missed-depeg rate and $\mathrm{FAR}_a$ is the severe false-alarm rate.

The benchmark also evaluates probabilistic and continuous predictions. 
Depeg probability is scored with Brier score, and calibration is assessed following modern neural-network calibration practice~\cite{guo2017calibration}.
Predicted maximum deviation is evaluated using mean absolute error and root mean squared error in basis points:
\[
\mathrm{Brier}_a=\frac{1}{N}\sum_{i=1}^{N}
\left(\widehat{\risk}_{a,i}-\mathbf{1}\{\ell_i=\texttt{stress}\}\right)^2,
\]
\[
\mathrm{MAE}^{\Delta}_a=\frac{1}{N}\sum_{i=1}^{N}
\lvert \widehat{\Delta}_{a,i}-\Delta_i\rvert,
\qquad
\mathrm{RMSE}^{\Delta}_a=
\sqrt{\frac{1}{N}\sum_{i=1}^{N}(\widehat{\Delta}_{a,i}-\Delta_i)^2}.
\]

\subsection{Cost, Reliability, and Reproducibility}

Since \textsc{StableEval Arena} evaluates agentic systems in a risk-sensitive setting, it records operational behavior in addition to prediction quality. 
For every agent-case run, the benchmark logs prompt tokens, completion tokens, total tokens, latency, estimated inference cost, cost per case, valid-JSON status, failure status, raw output, parsed output, model backend, execution mode, prompt version, and input-packet hash. 
These logs distinguish predictive errors from malformed outputs and execution failures.

For each agent-case run, operational cost is formalized as
\[
\Gamma_{a,i}=c_{\mathrm{in}}n^{\mathrm{in}}_{a,i}
+c_{\mathrm{out}}n^{\mathrm{out}}_{a,i},
\]
where $n^{\mathrm{in}}_{a,i}$ and $n^{\mathrm{out}}_{a,i}$ are prompt and completion tokens, and $c_{\mathrm{in}}$ and $c_{\mathrm{out}}$ are backend-specific token prices. 
Agent-level cost and latency are
\[
\overline{\Gamma}_a=\frac{1}{N}\sum_{i=1}^{N}\Gamma_{a,i},
\qquad
\overline{\tau}_a=\frac{1}{N}\sum_{i=1}^{N}\tau_{a,i}.
\]
Let $\nu_{a,i}=1$ indicate valid JSON output and $\omega_{a,i}=1$ indicate execution failure. 
The valid-output and failure rates are
\[
\rho_a=\frac{1}{N}\sum_{i=1}^{N}\nu_{a,i},
\qquad
\Omega_a=\frac{1}{N}\sum_{i=1}^{N}\omega_{a,i}.
\]

The 120-case validation artifacts are retained as historical validation outputs. 
The 507-case scaling study is additive: it uses the same prompt packets, backend, output schema, calibration rule, evaluator, and metric definitions, while writing agent-specific outputs, logs, metadata, and reports under the scaling artifact tree. 
Raw and calibrated predictions are preserved separately, and the canonical 507 leaderboard records execution-mode grouping for baselines, native/framework-level agents, and adapter-based -Claw extensions.

For reproducibility, the benchmark preserves case lists, prompt packets, prompt templates, leakage checks, calibration rules, execution metadata, raw predictions, calibrated predictions, cost logs, evaluation scripts, and versioned leaderboards. 
This release-oriented design follows representative benchmark and dataset-documentation practice in which public code, documented data artifacts, transparent protocols, and reproducible evaluation outputs are part of the contribution itself~\cite{erickson2025tabarena,li2025tsfmbench,gebru2021datasheets}.
The purpose is not only to rank agents, but to make agentic AI evaluation auditable: a system that produces valid low-cost outputs but misses rare stress events exposes a financial-risk failure mode rather than a deployment-ready financial monitor.

\vspace{-1.2em}
\section{Results}
\label{sec:results}

Table~\ref{tab:stableeval-main-results-compact} reports results for the two benchmark blocks defined above: \textsc{StableEval-120-Enriched} for rare-risk diagnostic validation, and \textsc{StableEval-507-Natural} for full-pool robustness and scaling.
To keep comparisons clear and consistent, both result blocks separate baselines, native/framework-level agents, and the adapter-based -Claw variants. 
We emphasize Macro-F1, balanced accuracy, stress recall, missed-depeg rate, valid-output reliability, and cost, since raw accuracy alone is insufficient under stablecoin class imbalance.

\vspace{-1em}
\subsection{120-Case Stress-Enriched Validation}
\label{subsec:results-120}


\textsc{StableEval-120-Enriched} evaluates system behavior under a stress-enriched distribution, with 45 watch and 21 stress cases providing sharper rare-risk diagnostic coverage than the natural 507-case pool.
Among baselines, historical persistence is strongest, with Macro-F1 of 0.285 and balanced accuracy of 0.300. 
The always-stable and simple logistic baselines both reach accuracy of 0.450, but only Macro-F1 of 0.207, demonstrating why accuracy alone is not an adequate metric for stable/watch/stress evaluation.

Among native/framework-level agents, Nanobot performs best under the calibrated first-pass protocol, with accuracy of 0.442, Macro-F1 of 0.351, and balanced accuracy of 0.393. 
LangGraph follows with Macro-F1 of 0.337, while Hermes AI and EvoAgentX obtain Macro-F1 scores of 0.321 and 0.272. 
For adapter-based -Claw group, NanoClaw obtains Macro-F1 of 0.328 and OpenClaw obtains 0.318. 
Overall, five of the six agentic systems exceed the historical-persistence baseline on Macro-F1, indicating that calibrated agentic outputs improve aggregate three-class classification in the stress-enriched setting.

Class-level results further qualify this aggregate improvement. 
Agent gains are concentrated more in stable/watch discrimination than in severe stress detection. 
Nanobot and OpenClaw obtain the highest calibrated first-pass agent stress recall, at 0.095, while historical persistence reaches 0.333.  
Thus, \textsc{StableEval-120-Enriched} implies agentic systems can improve aggregate warning classification, yet still struggle with the rarest and most consequential stress outcomes.
This distinction is central to the trustworthiness motivation of the benchmark.
The 120-case validation does not support a claim that current agents reliably solve stablecoin stress prediction; instead, it demonstrates that a unified agentic evaluation protocol can distinguish aggregate classification performance from stress reliability. 
Thus, the 120-case block remains useful for diagnosing stress behavior that is diluted in the natural 507-case distribution.

\vspace{-1em}
\subsection{507-Case Natural-Distribution Scaling}
\label{subsec:results-507}

\textsc{StableEval-507-Natural} applies the same protocol to the full natural-distribution pool, where 441 of 507 cases are stable and raw accuracy is therefore especially misleading.
The always-stable baseline reaches accuracy of 0.870, but has zero watch recall and zero stress recall. 
The strongest overall baseline is simple ML/logistic, with Macro-F1 of 0.4019, balanced accuracy of 0.4930, and stress recall of 0.3810. 
Historical persistence also remains strong, with Macro-F1 of 0.3867, balanced accuracy of 0.4500, stress recall of 0.3333, and missed-depeg rate of 0.6667.

Among the agentic systems, LangGraph is strongest on the calibrated 507 leaderboard, with Macro-F1 of 0.3044 and balanced accuracy of 0.4404, followed closely by NanoClaw adapter (Macro-F1 0.3041, balanced accuracy 0.4352) and Hermes AI (Macro-F1 0.3011, balanced accuracy 0.4328). 
Nanobot obtains Macro-F1 of 0.2874 and balanced accuracy of 0.4275, EvoAgentX obtains 0.2626 and 0.4138, and OpenClaw adapter obtains 0.2485 and 0.4032. 
Unlike in the 120-case stress-enriched validation, no agentic system exceeds the simple ML/logistic baseline on Macro-F1 in the natural-distribution setting.

The 507-case results also clarify the kind of behavior the agents learn under natural class balance. 
Most agentic systems remain reasonably sensitive to watch cases, but stress recall is still low. 
Nanobot and Hermes AI achieve the highest calibrated agent stress recall, at 0.0952, while LangGraph, EvoAgentX, OpenClaw, and NanoClaw each reach 0.0476. 
Thus, full-pool results confirms that \textsc{StableEval Arena} can collect valid JSON across heterogeneous systems on the 507-case pool, but rare-stress and sustained-depeg detection remain the main remaining limitation.

\definecolor{stableNavy}{HTML}{0F172A}
\definecolor{stableDark}{HTML}{334155}
\definecolor{stableGray}{HTML}{64748B}
\definecolor{stableBg}{HTML}{F8FAFC}

\definecolor{stableData}{HTML}{0284C7}
\definecolor{stableDataCard}{HTML}{E0F2FE}

\definecolor{stableProtocol}{HTML}{7C3AED}
\definecolor{stableProtocolCard}{HTML}{F3E8FF}

\definecolor{stableEval}{HTML}{0F766E}
\definecolor{stableEvalCard}{HTML}{DDF7F3}

\definecolor{stableAmber}{HTML}{D97706}
\definecolor{stableAmberCard}{HTML}{FEF3C7}

\definecolor{stableRisk}{HTML}{DB2777}
\definecolor{stableRiskCard}{HTML}{FCE7F3}

\definecolor{stableRule}{HTML}{0369A1}

\definecolor{stableGold}{HTML}{B45309}
\definecolor{stableSilver}{HTML}{64748B}
\definecolor{stableBronze}{HTML}{FB923C}

\providecommand{\best}[1]{}
\providecommand{\second}[1]{}
\providecommand{\third}[1]{}

\renewcommand{\best}[1]{\textbf{\textcolor{stableGold}{#1}}}
\renewcommand{\second}[1]{\textbf{\textcolor{stableSilver}{#1}}}
\renewcommand{\third}[1]{\textbf{\textcolor{stableBronze}{#1}}}

\newcommand{\stableSectionRowWide}[2]{%
\rowcolor{#1}
\multicolumn{11}{l}{\textbf{\textcolor{stableNavy}{#2}}}\\[-0.1em]
}

\begin{table*}[t]
\centering
\caption{Main results for Two Completed StableEval Benchmark Blocks}
\label{tab:stableeval-main-results-compact}
\vspace{-0.45em}
\scriptsize
\setlength{\tabcolsep}{2.7pt}
\renewcommand{\arraystretch}{0.92}
\arrayrulecolor{stableRule}
\begin{threeparttable}
\resizebox{\textwidth}{!}{%
\begin{tabular}{>{\raggedright\arraybackslash}p{1.35cm}
                >{\raggedright\arraybackslash}p{2.05cm}
                >{\raggedright\arraybackslash}p{2.05cm}
                rrrrrrrr}
\toprule[1.25pt]
\rowcolor{white}
\textbf{\textcolor{stableNavy}{Group}} &
\textbf{\textcolor{stableNavy}{System}} &
\textbf{\textcolor{stableNavy}{Execution}} &
\textbf{\textcolor{stableNavy}{Cases}} &
\textbf{\textcolor{stableNavy}{Acc. ($\uparrow$)}} &
\textbf{\textcolor{stableNavy}{Macro-F1 ($\uparrow$)}} &
\textbf{\textcolor{stableNavy}{Bal. Acc. ($\uparrow$)}} &
\textbf{\textcolor{stableNavy}{Stress R ($\uparrow$)}} &
\textbf{\textcolor{stableNavy}{MDR ($\downarrow$)}} &
\textbf{\textcolor{stableNavy}{Valid ($\uparrow$)}} &
\textbf{\textcolor{stableNavy}{Cost/case ($\downarrow$)}} \\
\midrule[0.85pt]

\stableSectionRowWide{stableProtocolCard}{\textsc{StableEval-120-Enriched}: 120-case stress-enriched validation, 54 stable / 45 watch / 21 stress}

Baseline & Always stable & deterministic & 120 & \best{0.450} & 0.207 & 0.333 & 0.000 & 1.000 & 1.00 & \$0 \\
Baseline & Historical persistence & deterministic & 120 & 0.292 & 0.285 & 0.300 & \best{0.333} & \best{0.667} & 1.00 & \$0 \\
Baseline & Threshold rule & deterministic & 120 & 0.283 & 0.278 & 0.295 & \best{0.333} & 1.000 & 1.00 & \$0 \\
Baseline & Simple ML/logistic & ML baseline & 120 & \best{0.450} & 0.207 & 0.333 & 0.000 & 1.000 & 1.00 & \$0 \\

\rowcolor{stableEvalCard}
Native & Nanobot & native/framework & 120 & 0.442 & \best{0.351} & \best{0.393} & 0.095 & 1.000 & 1.00 & \best{\$0.0029} \\
\rowcolor{stableBg}
Native & LangGraph & native/framework & 120 & 0.442 & \second{0.337} & 0.381 & 0.048 & 1.000 & 1.00 & \$0.0101 \\
Native & Hermes AI & native/framework & 120 & 0.425 & 0.321 & 0.369 & 0.048 & 1.000 & 1.00 & \best{\$0.0029} \\
\rowcolor{stableBg}
Native & EvoAgentX & native/framework & 120 & 0.392 & 0.272 & 0.348 & 0.048 & 1.000 & 1.00 & \$0.0051 \\

Adapter & OpenClaw & StableEval adapter & 120 & 0.408 & 0.318 & 0.368 & 0.095 & 1.000 & 1.00 & \$0.0052 \\
\rowcolor{stableBg}
Adapter & NanoClaw & StableEval adapter & 120 & 0.442 & \third{0.328} & \second{0.385} & 0.048 & 1.000 & 1.00 & \$0.0052 \\

\midrule[0.85pt]
\stableSectionRowWide{stableDataCard}{\textsc{StableEval-507-Natural}: 507-case natural-distribution scaling, 441 stable / 45 watch / 21 stress}

Baseline & Always stable & deterministic & 507 & \best{0.870} & 0.310 & 0.333 & 0.000 & 1.000 & 1.00 & \$0 \\
\rowcolor{stableRiskCard}
Baseline & Historical persistence & deterministic & 507 & 0.673 & \second{0.387} & 0.450 & 0.333 & \best{0.667} & 1.00 & \$0 \\
Baseline & Threshold rule & deterministic & 507 & 0.645 & 0.376 & 0.446 & 0.333 & 1.000 & 1.00 & \$0 \\
\rowcolor{stableEvalCard}
Baseline & Simple ML/logistic & ML baseline & 507 & 0.641 & \best{0.402} & \best{0.493} & \best{0.381} & 1.000 & 1.00 & \$0 \\

Native & Nanobot & native/framework & 507 & 0.394 & 0.287 & 0.428 & 0.095 & 1.000 & 1.00 & \best{\$0.0029} \\
\rowcolor{stableEvalCard}
Native & LangGraph & native/framework & 507 & 0.467 & \third{0.304} & \third{0.440} & 0.048 & 1.000 & 1.00 & \$0.0098 \\
Native & Hermes AI & native/framework & 507 & 0.408 & 0.301 & 0.433 & 0.095 & 1.000 & 1.00 & \best{\$0.0029} \\
\rowcolor{stableBg}
Native & EvoAgentX & native/framework & 507 & 0.381 & 0.263 & 0.414 & 0.048 & 1.000 & 1.00 & \$0.0050 \\

Adapter & OpenClaw & StableEval adapter & 507 & 0.353 & 0.248 & 0.403 & 0.048 & 1.000 & 1.00 & \$0.0050 \\
\rowcolor{stableBg}
Adapter & NanoClaw & StableEval adapter & 507 & 0.471 & 0.304 & 0.435 & 0.048 & 1.000 & 1.00 & \$0.0050 \\

\bottomrule[1.25pt]
\end{tabular}%
}
\end{threeparttable}

\vspace{-0.2em}
\noindent\parbox{\textwidth}{%
\footnotesize
\textbf{Notes.} Agent rows report calibrated outputs. ``Valid'' is the structured-output rate; for LLM-backed systems this is valid JSON rate. Stress R is recall on the enhanced stress class ($n=21$ per block; one case $=0.0476$). MDR is missed strict-depeg rate for sustained $\geq$100 bps depegs over 3h ($n=3$ per block; one case $=0.333$), computed from binary depeg probability and not equal to $1-\text{Stress R}$. Cost/case reports estimated model/API token cost only; deterministic and ML baselines have zero measured API cost. Accuracy is contextual; Macro-F1, balanced accuracy, Stress R, and MDR are the primary trustworthiness metrics. Full raw/calibrated rows and diagnostic artifacts are reported in the appendix.
}

\vspace{-0.7em}
\arrayrulecolor{black}
\end{table*}

\vspace{-1.5em}
\subsection{Cross Alignment and Stress Diagnostics}
\label{subsec:results-alignment-stress}

The 120-vs-507 alignment audits compare each \textsc{StableEval-120-Enriched} result with the corresponding \textsc{StableEval-507-Natural} run restricted to the same 120 case identifiers.
The restricted 507 results are generally close to or better than the 120 results, indicating that the scaling runs do not contradict the original stress-enriched validation. 
The lower or different full-pool metrics in \textsc{StableEval-507-Natural} mainly reflect the stable-dominant natural distribution rather than a protocol inconsistency.

The contrast between the two benchmark blocks is substantive rather than contradictory. 
\textsc{StableEval-120-Enriched} shows that calibrated agents can improve aggregate classification when watch and stress cases are concentrated. 
\textsc{StableEval-507-Natural} shows that under the full stable-dominant historical distribution, simple baselines remain strong and agentic systems do not yet dominate classical methods. 
Together, the two blocks provide a complete evaluation: one emphasizes stress sensitivity, while the other tests full-pool robustness and execution behavior.

The stress-case audit gives the main risk signal. 
There are 21 true stress cases in the 507-case pool. 
The four native-level agents and the two adapter-based -Claw variants miss 19 out of 21. 
Only 2 stress cases are caught by at least one system. 
No agentic system reduces missed-depeg below 1.0000; historical persistence is the only reported row with a lower missed-depeg rate, at 0.6667.
Hence, adding more agentic systems improves benchmark coverage and execution completeness, but does not materially resolve rare-stress detection. 
Thus, weak rare-stress detection is consistent across both evaluation blocks and remains the central limitation exposed by \textsc{StableEval Arena}.

\vspace{-1.5em}
\subsection{Calibration, Ablation, and Cost-Reliability Tradeoffs}
\label{subsec:results-ablation-cost}

Calibration improves the 120-case first-pass results by converting continuous risk estimates and maximum-deviation predictions into more useful stable/watch/stress labels. 
The best agent Macro-F1 rises from 0.301 in the raw setting to 0.351 after calibration, and the best agent stress recall rises from 0.000 to 0.095. 
This gain is modest but shows that part of the performance gap comes from label mapping rather than only from the underlying evidence packet.

The verifier layer produces a cost-sensitive tradeoff. 
It raises the best stress recall to 0.143, achieved by Nanobot and Hermes AI, but the best verifier-calibrated Macro-F1 is 0.349, slightly below the best calibrated first-pass Macro-F1 of 0.351. 
It also increases inference cost. 
For Nanobot, total cost rises from \$0.345 under calibrated first-pass inference to \$0.978 under verifier-calibrated inference. 
Thus, verifier-style auditing improves stress sensitivity, but the gain is limited and comes with a clear cost penalty.

The anonymization ablation probes whether results depend strongly on real stablecoin names or dates. 
The maximum label-change rate is 0.333, and mean probability changes range from 0.017 to 0.062. 
This indicates some identifier sensitivity, but not enough to suggest that the benchmark is driven primarily by memorized names or calendar effects, which we therefore treat as a reproducibility caveat rather than a protocol failure.

Executional reliability is consistently high. 
All main 120-case agent runs achieve valid JSON rate of 1.0 and failure rate of 0.0, and completed 507-case runs show that structured-output collection remains reliable at full-pool scale. 
This reliability result distincts from predictive performance: systems can follow the protocol and produce valid outputs, though rare-stress detection remains weak.

\vspace{-1em}
\subsection{Summary}
\label{subsec:results-summary}

The results show four main patterns. 
First, in the 120-case stress-enriched validation, calibrated agents improve aggregate Macro-F1 over simple baselines. 
Second, in the 507-case natural-distribution, the benchmark runs successfully across all cases, but simple ML/logistic remains stronger than all agentic systems on Macro-F1 and balanced accuracy. 
Third, 507 results show why accuracy alone is unreliable under stable-dominant class balance.
Fourth, rare-stress and sustained-depeg detection remain the main limitation: agents usually follow the output schema, but still miss most stress cases.

\vspace{-1em}
\section{Discussion and Analysis}
\label{sec:discussion}

The main result of \textsc{StableEval Arena} is a separation between procedural reliability and financial-risk reliability. 
The evaluated systems can execute a shared historical-replay protocol and return valid structured predictions across the full pool.
However, this reliability does not translate into dependable detection of rare peg stress or sustained depeg. 
\textsc{StableEval Arena} should therefore be read not as evidence that current agentic systems solve stablecoin stress prediction, but as evidence that shared agentic evaluation can expose a gap between protocol-following reliability and rare-event financial trustworthiness.

\vspace{-1em}
\subsection{Two-Block Evaluation}

The two benchmark blocks shown in Table~\ref{tab:stableeval-main-results-compact} answer complementary questions. 
\textsc{StableEval-120-Enriched} asks whether agents can reason about windows where watch and stress behavior is frequent enough to measure. 
By concentrating non-stable cases, it prevents stress behavior from being washed out by the stablecoin base rate. 
In this setting, calibrated agents improve Macro-F1 relative to the strongest simple baseline, indicating that structured evidence packets and calibration can support useful stable/watch/stress judgments. 
However, this result should be interpreted narrowly: low stress recall shows that better aggregate classification is not the same as reliable severe-risk detection.

\textsc{StableEval-507-Natural} asks whether the same protocol remains informative over the full pool under natural class balance. 
Because 441 of 507 cases are stable, accuracy becomes a weak indicator of risk sensitivity. 
The always-stable baseline makes this explicit: it achieves high accuracy while recovering no stress cases. 
The full-pool result also changes the comparison with baselines. 
Simple ML/logistic remains strongest by Macro-F1 and balanced accuracy, so agentic execution alone does not dominate classical methods once the natural stable majority is restored.

Together, the two-block design avoids two symmetric overreadings. 
The enriched block prevents the benchmark from concluding that systems are reliable simply because most cases are stable. 
The natural block prevents stress-enriched gains from being mistaken for deployment-like robustness. 
Thus, \textsc{StableEval Arena} should be interpreted less as a single leaderboard and more as an evaluation framework: shared prompts, backend, schema, calibration, metrics, and execution logs support reproducible diagnostic comparison across stress-enriched and natural-distribution settings~\cite{erickson2025tabarena,zhang2026predictionarena}.

\vspace{-1em}
\subsection{Reliability vs. Financial Trustworthiness}

Operationally, the systems are reliable; financially, stress detection remains weak.
The six evaluated systems complete the protocol with valid outputs across 507 cases, and the completed 507-case runs show that heterogeneous agents can be compared through one historical-replay interface. 
This matters because malformed outputs, missing predictions, runtime failures, and unlogged costs can otherwise make agent comparisons difficult to reproduce.

The negative finding is financial. 
Valid, low-cost JSON outputs are not equivalent to trustworthy monitoring.  
In the 507-case stress audit, all six systems miss 19 of 21 true stress cases. 
On the sustained-depeg audit, no agentic system in Table~\ref{tab:stableeval-main-results-compact} reduces missed-depeg rate below 1.0000.
A system that appears reliable by execution metrics can remain unreliable for event type that matters most. 
In financial monitoring, missed stress and false alarms often have asymmetric costs, so evaluation should not treat all classification errors as equally important.
For this reason, evaluation must report stress recall, missed-depeg rate, false alarms, calibration, and deviation error alongside accuracy, latency, cost, and valid-output rate.

This distinction aligns with agent-evaluation work emphasizing runtime behavior, diagnostics, and failure modes beyond final-answer accuracy~\cite{chen2026agent2rlbench}. 
For \textsc{StableEval Arena}, the primary failure mode is not inability to follow the output contract. 
It is the ability to follow the contract while still missing rare financial risk.

\vspace{-1em}
\subsection{Why Rare Stress Remains Difficult}

Rare-stress detection is difficult partly because the input evidence is intentionally bounded. 
The 30-day hourly lookback makes the benchmark reproducible and leakage-safe, but compact price and market evidence cannot fully represent reserves, redemption frictions, exchange depth, collateral composition, governance behavior, oracle conditions, cross-chain liquidity, or market confidence. 
This limitation is consistent with stablecoin research showing that peg stability depends on market design, governance, liquidity, and stakeholder context, not price history alone~\cite{zhang2025sokstablecoins,lyons2023stablecoins,arner2020stablecoins,klagesmundt2020stablecoins}.

The results also suggest conservative decision behavior under ambiguity. 
Since most windows remain near peg, agents often choose stable or watch unless the packet contains unusually strong stress evidence. 
That tendency can help aggregate performance in enriched 120-case block, but it is poorly aligned with high-consequence monitoring, where missing a rare stress event may be more costly than issuing some false alarms. 
Calibration and verifier layers can change the threshold between risk scores and labels, but they cannot recover signals that the packet does not contain.

This makes the benchmark's contribution precise. It does not show that current agents solve stablecoin risk prediction. 
It shows where their reliability stops: they can process bounded evidence and produce structured, reproducible assessments, but rare-stress detection still requires richer signals, more risk-sensitive decision rules, or calibration objectives that explicitly penalize missed depegs. 
This also points to future extensions using multimodal market evidence, stablecoin-arrangement infrastructure evidence, and oracle-network/interoperability evidence~\cite{fu2024dam,bis2022pfmistablecoins,cong2025oracle}.

\vspace{-1.2em}
\section{Industry Implications and Future Work}
\label{sec:industry-future}

\textsc{StableEval Arena} connects agentic AI evaluation to a broader industry question: whether agentic systems can become reliable monitoring layers for stablecoin-based payment infrastructure. 
\textsc{StableEval-120-Enriched} and \textsc{StableEval-507-Natural} show that agents can produce structured outputs reliably, yet weak rare-stress detection remains a barrier to automated payment execution. 
Thus, procedural reliability does not imply reliable financial-risk detection: an agent may satisfy interface requirements while still missing rare events that affect payment safety. 
This finding aligns with DeFi and stablecoin research that frames blockchain-based finance as programmable infrastructure while emphasizing smart-contract, governance, oracle, regulatory, interoperability, and stakeholder-specific risks~\cite{harvey2024international,zhang2025sokstablecoins}. 
We therefore highlight three industry implications as motivation for future extensions of StableEval.

The first industry implication concerns delegated authorization and interoperability for agentic and recurring payments. 
Web2 subscription systems rely on mature card, banking, and platform rails with established authorization, dispute, and refund mechanisms. 
By contrast, stablecoin-based recurring payments require users to delegate payment authority to applications, smart contracts, or agentic services~\cite{li2026stablecoinretailpayments}. 
Recent academic and policy work instead frames stablecoin payments as an unsettled design space involving authorization, interoperability, recourse, settlement, and risk allocation~\cite{li2026stablecoinretailpayments,wang2023exploringblockchainsinteroperability,cpmi2023stablecoinarrangements}.
However, these efforts remain competing emerging directions rather than settled standards or broadly adopted market practice.
Their significance is that they make user consent, authorization scope, spending limits, revocation, auditability, and liability concrete engineering requirements for agentic payment systems. 
\textsc{StableEval Arena} suggests that agentic monitors could help decide when to execute, pause, or reroute payments, but only if they can detect peg stress and settlement risk with higher stress recall than observed here.

The second industry implication concerns SLA design, clearing, and economic settlement finality. 
For stablecoins to support large-scale application payments, systems must specify when a payment is final, how reconciliation is performed, how failed or delayed settlement is escalated, and who bears the resulting loss. 
Regulatory discussions increasingly frame stablecoin arrangements as payment and financial-market infrastructure, applying concepts such as governance, risk management, settlement finality, money settlement, issuer reserves, and supervision~\cite{fsb2023globalstablecoins,cpmi2023stablecoinarrangements,bis2022pfmistablecoins,geniusAct2025,eu2023mica}.
These documents describe a regulatory target state, but operational settlement practice is not yet mature in deployed systems. 
In practice, on-chain confirmation alone does not establish economic finality: a stablecoin payment becomes economically final only when the recipient can use the funds, liquidity is available, compliance checks have cleared, reconciliation is complete, and responsibility for failure or delay is assigned~\cite{cpmi2023stablecoinarrangements,bis2022pfmistablecoins}. 
This gap is directly connected to \textsc{StableEval Arena}: our results show that an agent can produce valid JSON with zero runtime failures while still missing rare stress events, which in a payment SLA could lead to delayed settlement, failed reconciliation, or unsafe automatic execution. 
Future benchmarks should therefore evaluate not only next-window price classification, but also detection latency, reconciliation mismatch risk, settlement-delay risk, escalation behavior, and the cost of missed stress under explicit SLA rules.

The third industry implication concerns settlement pricing and tax accountability in automated stablecoin payments. 
Settlement pricing also creates accountability problems because stablecoin prices can diverge across venues, oracle feeds, and liquidity pools during market stress. 
Prior work on stablecoin depegging, market spillovers, and oracle risk shows that venue-specific prices and external data feeds can affect downstream liquidation, settlement, and payment outcomes~\cite{lee2025stablecoindepegging,perezriaza2025depegs,duley2023oracle}.
When prices diverge across venues, the system must decide whether the user, application, platform, exchange, or another intermediary absorbs the spread. 
Tax handling creates a parallel accountability problem: systems must determine the applicable jurisdiction, calculate fees and value-added taxes, and assign responsibility for remittance. 
These pricing and tax questions reflect a broader payment-design gap, since stablecoin arrangements differ from card networks in transaction lifecycle, recourse mechanisms, and risk allocation~\cite{li2026stablecoinretailpayments}. 
Hence, \textsc{StableEval Arena} can be extended with business-cost labels for false interruptions, unfavorable settlement, spread losses, underpayment, tax misclassification, and compliance disputes.

These implications suggest several future directions. 
First, future versions should broaden the asset universe and incorporate richer on-chain, off-chain, and unstructured evidence, including liquidity, redemption, exchange depth, oracle feeds, reserve attestations, market news, wallet-level flows, authorization mandates, revocation logs, audit trails, and settlement records. 
Second, evaluation should move from prediction quality alone to stakeholder-aware and risk-sensitive cost functions, including missed-stress penalties, detection-latency costs, user loss, developer revenue risk, platform compliance burden, settlement-price error, spread allocation, tax-calculation risk, and jurisdictional compliance risk~\cite{aldasoro2026impactstablecoins,li2026stablecoinretailpayments,zhang2025sokstablecoins}.
Third, future work should test multiple LLM backends, retain simple ML baselines as mandatory controls, add anonymized and counterfactual identifier checks, and evaluate multi-agent payment orchestration under explicit authorization, SLA, pricing, and tax constraints.
In this direction, \textsc{StableEval Arena} can evolve from a price-stability benchmark into a broader testbed for trustworthy agentic payment infrastructure.

\vspace{-0.5em}
\section{Limitations}
\label{sec:limitations}

We identify several limitations for \textsc{StableEval Arena}.
First, severe stablecoin stress cases are rare.
\textsc{StableEval-507-Natural} contains only 21 stress cases, while \textsc{StableEval-120-Enriched} intentionally concentrates them for diagnostic evaluation. 
Accordingly, stress recall and missed-depeg results should be read as risk warning signals rather than final statistical conclusions.
This limitation is structural: stablecoins usually remain near peg, while consequential failures occur in short, unusual windows.
The asset scope is also limited to USDT, USDC, and DAI, so future releases should test broader stablecoin designs and related payment assets.

Second, the benchmark prioritizes bounded, reproducible evidence packets over exhaustive real-world surveillance. 
The current packets include price, volume, market-structure, and contextual signals available in the frozen lookback window, but they do not include every high-cost or real-time risk signal that might matter in practice. 
Future extensions will add reserve attestations, redemption flows, cross-chain liquidity, DeFi liquidation activity, oracle feeds, regulatory events, and news while preserving leakage-safe historical replay.
Additionally, using \texttt{qwen/qwen-2.5-72b-instruct} via OpenRouter improves comparability across agentic systems, but it does not test backend generality. 
Similarly, OpenClaw and NanoClaw are reported as StableEval adapter-based compatibility extensions rather than native CMDOP executions.

Finally, residual parametric-memory effects cannot be fully ruled out because some packets retain real stablecoin names and dates, even though future labels are hidden and anonymization checks are included. 
Future releases will broaden anonymized and counterfactual evaluation and strengthen dataset cards and artifact manifests, following benchmark and datasheet practices~\cite{gebru2021datasheets,erickson2025tabarena}.
These boundaries define the current benchmark setting and paths for extending trustworthy agentic AI evaluation in stablecoin risk monitoring.

\section{Conclusion}
\label{sec:conclusion}

We introduce \textsc{StableEval Arena}, a cost-aware agentic benchmark for stablecoin price-stability risk assessment. 
Each case provides a 30-day evidence window and requires structured next-7-day predictions of stable/watch/stress status, depeg probability, maximum peg deviation, direction, confidence, and rationale. 
The benchmark compares agentic systems with simple baselines under a unified protocol and evaluates prediction quality, stress recall, missed-depeg risk, calibration, valid-output rate, latency, token usage, and inference cost.
The completed experiments combine a 120-case stress-enriched validation for rare-risk diagnosis with a 507-case natural-distribution scaling study for full-pool robustness and operational evaluation. 
Rather than showing current agents can act as deployment-ready stablecoin risk monitors, our results show that a leakage-safe, cost-aware agentic benchmark can reveal the gap between valid execution and trustworthy rare-event financial detection.
Across both settings, systems produce valid structured outputs, but rare-stress and sustained-depeg detection remain weak.
This gap between execution reliability and financial-risk detection is the central finding of \textsc{StableEval Arena}. 
Future work should broaden evidence packets and asset coverage, test backend sensitivity and more robust native agent execution, and evaluate richer payment, settlement, and stablecoin-risk monitoring scenarios.

\bibliography{reference}
\bibliographystyle{ACM-Reference-Format}

\clearpage
\onecolumn
%

\appendix

\begingroup
\raggedbottom
\emergencystretch=2em

\providecommand{\Coins}{\mathscr{C}}
\providecommand{\Agents}{\mathscr{A}}
\providecommand{\Cases}{\mathscr{D}}
\providecommand{\Look}{\mathscr{L}}
\providecommand{\Horizon}{\mathscr{H}}
\providecommand{\Labels}{\mathscr{Y}}
\providecommand{\Features}{\bm{x}}
\providecommand{\depeg}{\delta}
\providecommand{\risk}{\varpi}
\providecommand{\conf}{\kappa}

\newcommand{\StableEvalCompactTable}{%
  \footnotesize
  \setlength{\tabcolsep}{3.5pt}%
  \renewcommand{\arraystretch}{1.12}%
}

\newcommand{\StableEvalTinyTable}{%
  \scriptsize
  \setlength{\tabcolsep}{2.8pt}%
  \renewcommand{\arraystretch}{1.10}%
}

\newcommand{\StableEvalRoomyTable}{%
  \small
  \setlength{\tabcolsep}{4.2pt}%
  \renewcommand{\arraystretch}{1.14}%
}

\newcommand{\StableEvalArtifactTable}{%
  \scriptsize
  \setlength{\tabcolsep}{3pt}%
  \renewcommand{\arraystretch}{1.10}%
}

\newcommand{\StableEvalLandscapeStart}{%
  \clearpage
  \begin{landscape}
  \StableEvalCompactTable
}

\newcommand{\StableEvalLandscapeEnd}{%
  \end{landscape}
  \clearpage
}

\newcommand{\StableEvalPathCell}[1]{%
  \begin{tabular}[t]{@{}l@{}}#1\end{tabular}%
}

\definecolor{stableevalBlue}{HTML}{9EC3E6}
\definecolor{stableevalOrange}{HTML}{F1B88C}
\definecolor{stableevalGreen}{HTML}{8FD19E}
\definecolor{stableevalPurple}{HTML}{B8A7E6}
\definecolor{stableevalTeal}{HTML}{7FC7C0}
\definecolor{stableevalRed}{HTML}{D88484}
\definecolor{stableevalGray}{HTML}{F3F4F6}

\setlength{\textfloatsep}{8pt plus 2pt minus 2pt}
\setlength{\floatsep}{8pt plus 2pt minus 2pt}
\setlength{\intextsep}{8pt plus 2pt minus 2pt}
\setlength{\abovecaptionskip}{4pt}
\setlength{\belowcaptionskip}{2pt}

\section{Shared Protocol, Data, and Execution Metadata}
\label{app:detailed-results}
\label{app:shared-protocol}

This appendix provides the audit trail behind two implementation-result blocks. \textsc{StableEval-120-Enriched} is the frozen 120-case stress-enriched validation block used for the main six-system comparison. \textsc{StableEval-507-Natural} is the full 507-case natural-distribution scaling block. Both blocks use 30-day lookback windows, hidden 7-day prediction horizons, the revised risk-sensitive prompt packet, \texttt{qwen/qwen-2.5-72b-instruct} through OpenRouter, temperature 0, strict JSON outputs, and the same evaluator logic. The 507-case block is additive: it tests scaling and natural-distribution behavior, while the 120-case block remains the primary rare-risk validation because it is stress-enriched.

\subsection{Glossary of Notation and Key Terms}
\label{app:glossary}

{\StableEvalRoomyTable
\begin{longtable}{@{}p{0.11\linewidth}p{0.34\linewidth}p{0.11\linewidth}p{0.34\linewidth}@{}}
\caption{Glossary of key terms, abbreviations, and mathematical notation used in the appendix.}
\label{tab:glossary}\\
\toprule
\textbf{Term} & \textbf{Definition} & \textbf{Term} & \textbf{Definition} \\
\midrule
\endfirsthead
\toprule
\textbf{Term} & \textbf{Definition} & \textbf{Term} & \textbf{Definition} \\
\midrule
\endhead
\bottomrule
\endlastfoot
$\Cases$ & Benchmark case set & $\xi_i$ & One historical-replay case \\
$\Coins$ & Target stablecoin universe & $\chi_i$ & Target stablecoin in case $i$ \\
$t_i$ & Observation time & $\Look_i$ & Frozen 30-day lookback window \\
$\Horizon_i$ & Hidden 7-day future horizon & $\Features_{\chi,\tau}$ & Feature vector at hour $\tau$ \\
$p_{\chi,\tau}$ & Hourly close price & $\depeg_{\chi,\tau}$ & Peg deviation in basis points \\
$\Delta_i$ & Maximum hidden-horizon deviation & $z_i$ & Strict sustained-depeg indicator \\
$\Labels$ & \texttt{stable}/\texttt{watch}/\texttt{stress} label space & $\ell_i$ & Hidden evaluator label \\
$\Agents$ & Evaluated agent set & $f_{\theta_a}$ & Agent prediction function \\
$\widehat{s}_{a,i}$ & Predicted stability label & $\widehat{\risk}_{a,i}$ & Predicted depeg-risk probability \\
$\widehat{\Delta}_{a,i}$ & Predicted max deviation & $\widehat{\phi}_{a,i}$ & Predicted deviation direction \\
$\widehat{\conf}_{a,i}$ & Agent confidence score & $\Gamma_{a,i}$ & Estimated inference cost \\
$\tau_{a,i}$ & Latency & $\rho_a$ & Valid-output rate \\
$\Omega_a$ & Failure rate & MDR/FAR & Missed-depeg and false-alarm rates \\
\end{longtable}
}

\subsection{Dataset, Labels, and Case Construction}
\label{app:dataset-labels}

This section defines the shared benchmark data used by both result blocks. Table~\ref{tab:dataset-construction} summarizes the conversion from hourly stablecoin market observations into historical-replay cases. The same source universe, 30-day lookback, 7-day hidden horizon, and stable/watch/stress label definitions are used for both blocks. Hidden future-horizon labels are used only for case selection and evaluation; they are never included in the agent-facing prompt packet.

{\StableEvalRoomyTable
\begin{longtable}{@{}p{0.24\linewidth}p{0.68\linewidth}@{}}
\caption{Dataset and case-construction summary for the shared StableEval benchmark protocol.}
\label{tab:dataset-construction}\\
\toprule
\textbf{Component} & \textbf{Value} \\
\midrule
\endfirsthead
\toprule
\textbf{Component} & \textbf{Value} \\
\midrule
\endhead
\bottomrule
\endlastfoot
Stablecoins & USDT, USDC, DAI \\
Context assets & BTC, ETH \\
Source date range & 2023-01-01 to 2026-05-01 \\
Frequency & Hourly price/volume; daily market-structure metadata joined by date \\
Lookback & 30 days \\
Prediction horizon & 7 days \\
Stride & 7 days \\
Full frozen case pool & 507 historical-replay cases \\
Strict sustained-depeg cases & 3 in the full frozen pool \\
Implementation result blocks & \textsc{StableEval-120-Enriched}; \textsc{StableEval-507-Natural} \\
\end{longtable}
}

\newpage
{\StableEvalRoomyTable
\begin{longtable}{@{}p{0.22\linewidth}p{0.70\linewidth}@{}}
\caption{Stable/watch/stress and depeg label definitions used by the evaluator.}
\label{tab:label-definitions}\\
\toprule
\textbf{Label or quantity} & \textbf{Definition} \\
\midrule
\endfirsthead
\toprule
\textbf{Label or quantity} & \textbf{Definition} \\
\midrule
\endhead
\bottomrule
\endlastfoot
Deviation bps & $\lvert\text{hourly close}-1.0\rvert \times 10000$ \\
Stable & No meaningful next-7-day instability under the enhanced label rule. \\
Watch & Mild but meaningful next-7-day instability, including any hourly close deviation at or above 50 bps absent strict sustained depeg. \\
Stress & Large, persistent, or operationally meaningful instability, including strict sustained depeg, 100 bps close deviation, repeated 50 bps hours, or high average absolute deviation. \\
Strict sustained depeg & Absolute hourly close deviation at or above 100 bps for at least three consecutive hours. \\
\end{longtable}
}

\subsection{Agentic Systems and Execution Metadata}
\label{app:agent-metadata}

\textsc{StableEval Arena} compares six platforms under a common model backend and shared prompt/evaluator protocol. Nanobot, LangGraph, Hermes AI, and EvoAgentX are native or framework-level agent comparisons. OpenClaw and NanoClaw are included as StableEval adapter executions. Because all LLM-backed systems use the same backend, the comparison measures platform-mediated execution, structured-output behavior, cost, latency, and reliability under a shared interface rather than independent foundation-model ability.

{\StableEvalCompactTable
\begin{longtable}{@{}p{0.10\linewidth}p{0.17\linewidth}p{0.10\linewidth}p{0.24\linewidth}p{0.31\linewidth}@{}}
\caption{StableEval Arena agent platform metadata.}
\label{tab:agent-platform-metadata}\\
\toprule
\textbf{Platform} & \textbf{Package} & \textbf{Native runtime} & \textbf{Execution mode} & \textbf{Caveat} \\
\midrule
\endfirsthead
\toprule
\textbf{Platform} & \textbf{Package} & \textbf{Native runtime} & \textbf{Execution mode} & \textbf{Caveat} \\
\midrule
\endhead
\bottomrule
\endlastfoot
Nanobot & nanobot-ai 0.2.0 & True & Nanobot SDK OpenRouter agent loop & Real local Nanobot SDK framework path executed with OpenRouter LLM backend. \\
LangGraph & langgraph 1.2.0 & True & Two-node StateGraph with OpenRouter LLM & Real local LangGraph StateGraph framework path executed with OpenRouter LLM backend. \\
Hermes AI & hermes-ai 0.3.20 & True & Hermes Agent with LlamaIndex/OpenRouter & Real local Hermes Agent framework path executed with OpenRouter LLM backend. \\
EvoAgentX & evoagentx 0.1.0 & True & EvoAgentX OpenRouter adapter & Real local EvoAgentX OpenRouter-backed workflow executed. \\
OpenClaw & openclaw 2026.3.20 & False & StableEval adapter & CMDOP local runtime was not available; StableEval used an LLM-backed adapter, not native CMDOP execution. \\
NanoClaw & nanoclaw 2026.3.20 & False & StableEval adapter & CMDOP local runtime was not available; StableEval used an LLM-backed adapter, not native CMDOP execution. \\
\end{longtable}
}

\subsection{Computing Environment and Resource Configuration}
\label{app:computing-environment}

StableEval primarily uses API-based LLM inference through OpenRouter. Local hardware affects orchestration, prompt-packet construction, logging, JSON parsing, calibration, scoring, analysis, and figure/table generation, while the LLM inference itself is served remotely.

{\StableEvalRoomyTable
\begin{longtable}{@{}p{0.27\linewidth}p{0.65\linewidth}@{}}
\caption{Execution-environment facts that are recorded or directly supported by the release artifacts.}
\label{tab:computing-environment}\\
\toprule
\textbf{Field} & \textbf{Recorded value or reproducibility note} \\
\midrule
\endfirsthead
\toprule
\textbf{Field} & \textbf{Recorded value or reproducibility note} \\
\midrule
\endhead
\bottomrule
\endlastfoot
Execution setting & API-backed LLM inference plus local orchestration and evaluation. \\
LLM backend & \texttt{qwen/qwen-2.5-72b-instruct} through OpenRouter. \\
Decoding temperature & 0. \\
Web browsing & Disabled in the reported protocol. \\
Local GPU/TPU use & No local GPU/TPU is required for saved-output recomputation; LLM inference is remote. \\
Local computation responsibilities & Build prompt packets, call platform wrappers, log outputs, parse strict JSON, apply calibration, compute metrics, and generate tables/figures. \\
Package/runtime metadata & \StableEvalPathCell{\texttt{results/stableeval\_120\_enriched/}\\\texttt{agent\_execution\_metadata.csv}} \\
Backend metadata & \StableEvalPathCell{\texttt{results/stableeval\_120\_enriched/}\\\texttt{model\_backend.json}\\\texttt{results/stableeval\_507\_natural/}\\\texttt{metadata/model\_backend.json}\\\texttt{protocols/}\\\texttt{stableeval\_507\_natural\_protocol.json}} \\
\end{longtable}
}

\section{StableEval-120-Enriched Validation Results}
\label{app:block-120-results}
\label{app:stableeval-120}

The 120-case stress-enriched validation block contains 54 stable cases, 45 watch cases, and 21 stress cases. It includes all enhanced stress cases and all strict sustained-depeg cases available in the frozen pool, together with watch cases, near-stress stable controls, and natural stable controls. This composition makes the 120-case block the primary comparison for warning quality and rare-event trustworthiness.

\subsection{120-Case Leaderboard-Style Summary Figures}
\label{app:leaderboard-figures}

\definecolor{stableNavy}{HTML}{0F172A}
\definecolor{stableDark}{HTML}{334155}
\definecolor{stableGray}{HTML}{64748B}
\definecolor{stableBg}{HTML}{F8FAFC}

\definecolor{stableData}{HTML}{38BDF8}
\definecolor{stableDataPanel}{HTML}{F0F9FF}
\definecolor{stableDataCard}{HTML}{E0F2FE}

\definecolor{stableProtocol}{HTML}{A78BFA}
\definecolor{stableProtocolPanel}{HTML}{FAF5FF}
\definecolor{stableProtocolCard}{HTML}{F3E8FF}

\definecolor{stableEval}{HTML}{2DD4BF}
\definecolor{stableEvalPanel}{HTML}{F0FDFA}
\definecolor{stableEvalCard}{HTML}{CCFBF1}

\definecolor{stableAmber}{HTML}{FACC15}
\definecolor{stableAmberCard}{HTML}{FEF9C3}

\definecolor{stableRisk}{HTML}{F472B6}
\definecolor{stableRiskCard}{HTML}{FDF2F8}

The following figures summarize the main 120-case validation patterns before the numeric tables. They are drawn from the reported validation metrics and use the StableEval color palette used in the main paper.

\begin{figure}[!htbp]
\centering
\begin{tikzpicture}
\begin{axis}[
    xbar,
    width=0.90\linewidth,
    height=0.38\linewidth,
    xmin=0,
    xmax=0.48,
    bar width=2.8pt,
    enlarge y limits=0.065,
    symbolic y coords={
        Simple ML,
        Always Stable,
        Threshold Rule,
        Historical Persistence,
        EvoAgentX,
        OpenClaw,
        Hermes AI,
        NanoClaw,
        LangGraph,
        Nanobot
    },
    ytick=data,
    yticklabel style={font=\scriptsize, text width=2.7cm, align=right},
    xtick={0,0.1,0.2,0.3,0.4},
    xlabel={Score},
    xlabel style={font=\small},
    tick label style={font=\scriptsize},
    xmajorgrids=true,
    grid style={draw=stableGray!22},
    axis line style={draw=stableGray!45},
    legend style={
        draw=stableGray!35,
        fill=white,
        font=\scriptsize,
        at={(0.5,1.04)},
        anchor=south,
        legend columns=3
    },
    extra x ticks={0.2854233827638083},
    extra x tick labels={Hist. persistence},
    extra x tick style={
        tick label style={font=\scriptsize, rotate=90},
        grid=major,
        major grid style={draw=stableNavy, densely dashed, line width=0.8pt}
    },
]
\addplot+[draw=stableData!80!black, fill=stableData!45] coordinates {
    (0.2068965517241379,Simple ML)
    (0.2068965517241379,Always Stable)
    (0.2779405113470083,Threshold Rule)
    (0.2854233827638083,Historical Persistence)
    (0.2719261633666122,EvoAgentX)
    (0.3175626175626175,OpenClaw)
    (0.3213489409141584,Hermes AI)
    (0.3277521140981951,NanoClaw)
    (0.3372427983539094,LangGraph)
    (0.3505430242272347,Nanobot)
};
\addplot+[draw=stableProtocol!80!black, fill=stableProtocol!45] coordinates {
    (0.3333333333333333,Simple ML)
    (0.3333333333333333,Always Stable)
    (0.2950617283950617,Threshold Rule)
    (0.3000000000000000,Historical Persistence)
    (0.3479717813051146,EvoAgentX)
    (0.3675485008818341,OpenClaw)
    (0.3689594356261023,Hermes AI)
    (0.3850088183421516,NanoClaw)
    (0.3813051146384480,LangGraph)
    (0.3934744268077601,Nanobot)
};
\addplot+[draw=stableRisk!80!black, fill=stableRisk!45] coordinates {
    (0.0000000000000000,Simple ML)
    (0.0000000000000000,Always Stable)
    (0.3333333333333333,Threshold Rule)
    (0.3333333333333333,Historical Persistence)
    (0.0476190476190476,EvoAgentX)
    (0.0952380952380952,OpenClaw)
    (0.0476190476190476,Hermes AI)
    (0.0476190476190476,NanoClaw)
    (0.0476190476190476,LangGraph)
    (0.0952380952380952,Nanobot)
};
\legend{Macro-F1, Balanced accuracy, Stress recall}
\end{axis}
\end{tikzpicture}
\caption{Validation leaderboard on the 120-case stress-enriched benchmark. Bars report Macro-F1, balanced accuracy, and stress-class recall from the calibrated first-pass protocol. The dashed reference line marks the Macro-F1 of the strongest non-agent baseline, historical persistence.}
\label{fig:validation-leaderboard}
\end{figure}
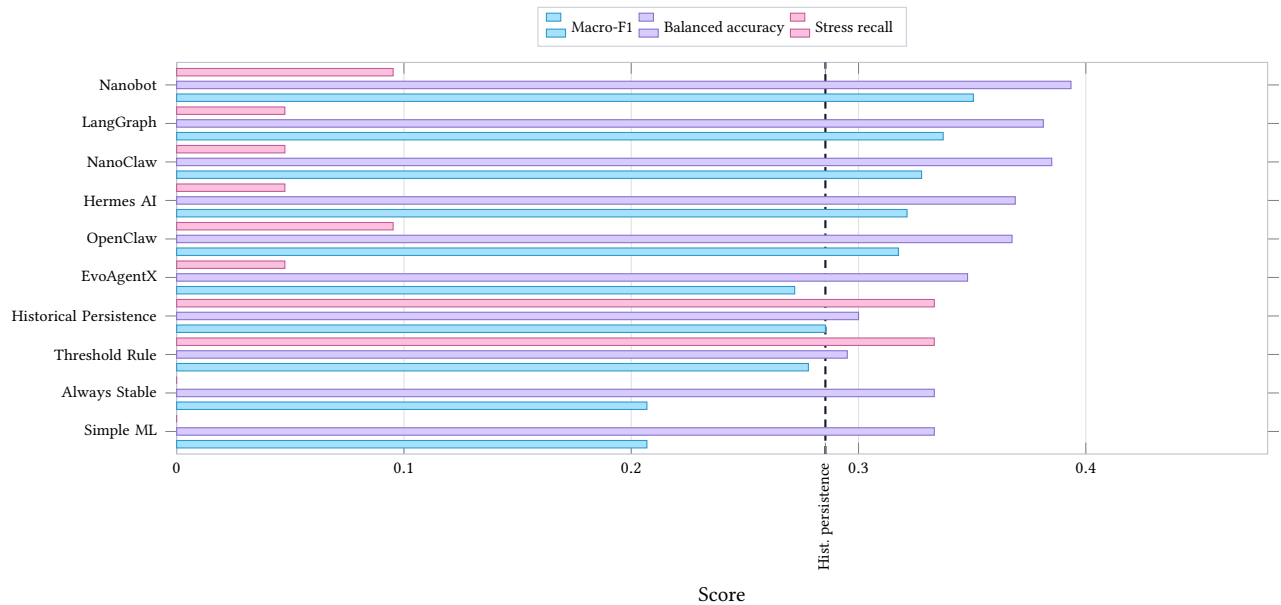

\begin{figure}[!htbp]
\centering
\begin{tikzpicture}
\begin{groupplot}[
    group style={group size=2 by 1, horizontal sep=1.05cm},
    width=0.45\linewidth,
    height=0.27\linewidth,
    tick label style={font=\scriptsize},
    label style={font=\small},
    axis line style={draw=stableGray!45},
    ymajorgrids=true,
    grid style={draw=stableGray!22},
]
\nextgroupplot[
    ybar,
    ymin=0,
    ymax=0.43,
    bar width=5pt,
    ylabel={Best score},
    symbolic x coords={BaseRaw,BaseCalib,VerifierRaw,VerifierCalib},
    xtick=data,
    xticklabels={Raw, Calib., Verifier raw, Verifier calib.},
    x tick label style={rotate=18, anchor=east},
    legend style={
        draw=stableGray!35,
        fill=white,
        font=\scriptsize,
        at={(0.5,1.10)},
        anchor=south,
        legend columns=3
    },
]
\addplot+[draw=stableData!80!black, fill=stableData!45] coordinates {
    (BaseRaw,0.3008811519449817)
    (BaseCalib,0.3505430242272347)
    (VerifierRaw,0.2721910915913642)
    (VerifierCalib,0.3491652965337176)
};
\addplot+[draw=stableProtocol!80!black, fill=stableProtocol!45] coordinates {
    (BaseRaw,0.3679012345679012)
    (BaseCalib,0.3934744268077601)
    (VerifierRaw,0.3504409171075838)
    (VerifierCalib,0.3883597883597883)
};
\addplot+[draw=stableRisk!80!black, fill=stableRisk!45] coordinates {
    (BaseRaw,0.0000000000000000)
    (BaseCalib,0.0952380952380952)
    (VerifierRaw,0.0952380952380952)
    (VerifierCalib,0.1428571428571428)
};
\legend{Macro-F1, Balanced accuracy, Stress recall}

\nextgroupplot[
    ybar,
    ymin=0,
    ymax=1.45,
    bar width=13pt,
    ylabel={Mean total cost (USD)},
    symbolic x coords={BaseRaw,BaseCalib,VerifierRaw,VerifierCalib},
    xtick=data,
    xticklabels={Raw, Calib., Verifier raw, Verifier calib.},
    x tick label style={rotate=18, anchor=east},
]
\addplot+[draw=stableAmber!80!black, fill=stableAmber!55] coordinates {
    (BaseRaw,0.6281394066666667)
    (BaseCalib,0.6281394066666667)
    (VerifierRaw,1.2596299266666666)
    (VerifierCalib,1.2596299266666666)
};
\node[font=\scriptsize, align=center, text=stableNavy] at (axis cs:VerifierRaw,1.36) {verifier\\cost doubles};
\end{groupplot}
\end{tikzpicture}
\caption{120-case ablation summary across raw first-pass, calibrated first-pass, and verifier protocols. Calibration improves the best aggregate scores. The verifier increases best stress recall but approximately doubles mean total cost.}
\label{fig:validation-ablation}
\end{figure}
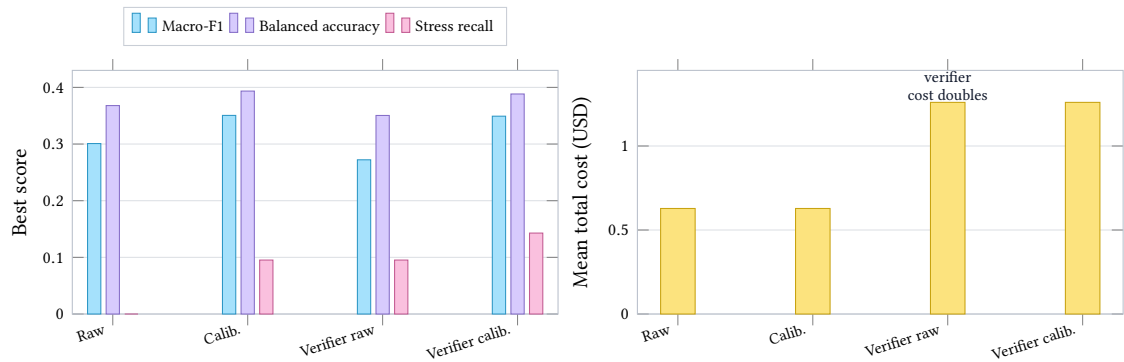

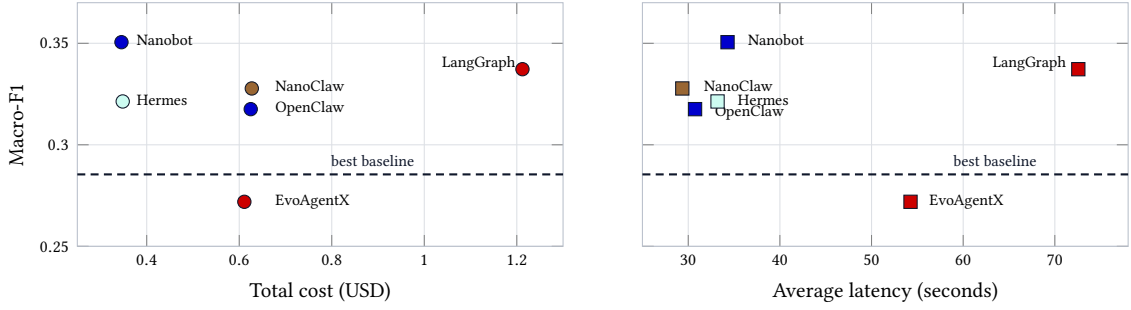
\begin{figure}[!htbp]
\centering
\begin{tikzpicture}
\begin{groupplot}[
    group style={group size=2 by 1, horizontal sep=1.05cm},
    width=0.45\linewidth,
    height=0.27\linewidth,
    ymin=0.25,
    ymax=0.37,
    tick label style={font=\scriptsize},
    label style={font=\small},
    axis line style={draw=stableGray!45},
    xmajorgrids=true,
    ymajorgrids=true,
    grid style={draw=stableGray!22},
]
\nextgroupplot[
    xmin=0.25,
    xmax=1.30,
    xlabel={Total cost (USD)},
    ylabel={Macro-F1},
]
\addplot[draw=stableNavy, densely dashed, line width=0.8pt, forget plot] coordinates {
    (0.25,0.2854233827638083)
    (1.30,0.2854233827638083)
};
\node[font=\scriptsize, anchor=south west, text=stableNavy] at (axis cs:0.78,0.2854233827638083) {best baseline};
\addplot+[only marks, mark=*, mark size=2.5pt, draw=stableNavy, fill=stableAmber] coordinates {(0.34541426,0.3505430242272347)};
\addplot+[only marks, mark=*, mark size=2.5pt, draw=stableNavy, fill=stableGray] coordinates {(1.212237,0.3372427983539094)};
\addplot+[only marks, mark=*, mark size=2.5pt, draw=stableNavy, fill=stableEval] coordinates {(0.62717796,0.3277521140981951)};
\addplot+[only marks, mark=*, mark size=2.5pt, draw=stableNavy, fill=stableEvalCard] coordinates {(0.34811488,0.3213489409141584)};
\addplot+[only marks, mark=*, mark size=2.5pt, draw=stableNavy, fill=stableData] coordinates {(0.6248405000000001,0.3175626175626175)};
\addplot+[only marks, mark=*, mark size=2.5pt, draw=stableNavy, fill=stableProtocol] coordinates {(0.61105184,0.2719261633666122)};
\node[font=\scriptsize, anchor=west] at (axis cs:0.36,0.352) {Nanobot};
\node[font=\scriptsize, anchor=west] at (axis cs:1.02,0.340) {LangGraph};
\node[font=\scriptsize, anchor=west] at (axis cs:0.66,0.329) {NanoClaw};
\node[font=\scriptsize, anchor=west] at (axis cs:0.36,0.322) {Hermes};
\node[font=\scriptsize, anchor=west] at (axis cs:0.66,0.318) {OpenClaw};
\node[font=\scriptsize, anchor=west] at (axis cs:0.66,0.272) {EvoAgentX};

\nextgroupplot[
    xmin=25,
    xmax=78,
    xlabel={Average latency (seconds)},
    yticklabels={},
]
\addplot[draw=stableNavy, densely dashed, line width=0.8pt, forget plot] coordinates {
    (25,0.2854233827638083)
    (78,0.2854233827638083)
};
\node[font=\scriptsize, anchor=south west, text=stableNavy] at (axis cs:58,0.2854233827638083) {best baseline};
\addplot+[only marks, mark=square*, mark size=2.5pt, draw=stableNavy, fill=stableAmber] coordinates {(34.286870933749015,0.3505430242272347)};
\addplot+[only marks, mark=square*, mark size=2.5pt, draw=stableNavy, fill=stableGray] coordinates {(72.56169760242143,0.3372427983539094)};
\addplot+[only marks, mark=square*, mark size=2.5pt, draw=stableNavy, fill=stableEval] coordinates {(29.35095737848411,0.3277521140981951)};
\addplot+[only marks, mark=square*, mark size=2.5pt, draw=stableNavy, fill=stableEvalCard] coordinates {(33.197878329150264,0.3213489409141584)};
\addplot+[only marks, mark=square*, mark size=2.5pt, draw=stableNavy, fill=stableData] coordinates {(30.742086340675208,0.3175626175626175)};
\addplot+[only marks, mark=square*, mark size=2.5pt, draw=stableNavy, fill=stableProtocol] coordinates {(54.25374271872473,0.2719261633666122)};
\node[font=\scriptsize, anchor=west] at (axis cs:35.6,0.352) {Nanobot};
\node[font=\scriptsize, anchor=west] at (axis cs:62.3,0.340) {LangGraph};
\node[font=\scriptsize, anchor=west] at (axis cs:30.8,0.329) {NanoClaw};
\node[font=\scriptsize, anchor=west] at (axis cs:34.5,0.322) {Hermes};
\node[font=\scriptsize, anchor=west] at (axis cs:32.0,0.316) {OpenClaw};
\node[font=\scriptsize, anchor=west] at (axis cs:55.4,0.272) {EvoAgentX};
\end{groupplot}
\end{tikzpicture}
\caption{Cost-aware 120-case performance tradeoffs for the six agentic systems. Nanobot has the strongest Macro-F1 at low total cost, while LangGraph is competitive but more expensive and slower under the frozen validation protocol.}
\label{fig:cost-performance-tradeoff}
\end{figure}

\newpage
\subsection{120-Case Leaderboard Tables}
\label{app:validation-tables}

Tables~\ref{tab:validation-120} and~\ref{tab:validation-120-ops} provide the compact calibrated 120-case result. The wider cross-variant matrices are retained at the end of this section because they are the most direct way to audit raw, calibrated, baseline-only, and verifier variants from the saved outputs.

{\StableEvalCompactTable
\begin{longtable}{@{}p{0.18\linewidth}p{0.09\linewidth}rrrrrr@{}}
\caption{120-case calibrated validation summary: predictive and risk metrics.}
\label{tab:validation-120}
\label{tab:agent-baseline}\\
\toprule
\textbf{System} & \textbf{Type} & \textbf{Acc.} & \textbf{Macro-F1} & \textbf{BalAcc} & \textbf{Stress R} & \textbf{MDR} & \textbf{FAR} \\
\midrule
\endfirsthead
\toprule
\textbf{System} & \textbf{Type} & \textbf{Acc.} & \textbf{Macro-F1} & \textbf{BalAcc} & \textbf{Stress R} & \textbf{MDR} & \textbf{FAR} \\
\midrule
\endhead
\bottomrule
\endlastfoot
Nanobot & agent & 0.442 & 0.351 & 0.393 & 0.095 & 1.000 & 0.009 \\
LangGraph & agent & 0.442 & 0.337 & 0.381 & 0.048 & 1.000 & 0.009 \\
NanoClaw & agent & 0.442 & 0.328 & 0.385 & 0.048 & 1.000 & 0.017 \\
Hermes AI & agent & 0.425 & 0.321 & 0.369 & 0.048 & 1.000 & 0.017 \\
OpenClaw & agent & 0.408 & 0.318 & 0.368 & 0.095 & 1.000 & 0.017 \\
EvoAgentX & agent & 0.392 & 0.272 & 0.348 & 0.048 & 1.000 & 0.017 \\
Historical persistence & baseline & 0.292 & 0.285 & 0.300 & 0.333 & 0.667 & 0.376 \\
Threshold rule & baseline & 0.283 & 0.278 & 0.295 & 0.333 & 1.000 & 0.282 \\
Always stable & baseline & 0.450 & 0.207 & 0.333 & 0.000 & 1.000 & 0.000 \\
Simple ML logistic & baseline & 0.450 & 0.207 & 0.333 & 0.000 & 1.000 & 0.000 \\
\end{longtable}
}

{\StableEvalCompactTable
\begin{longtable}{@{}p{0.17\linewidth}rrrrrrr@{}}
\caption{120-case calibrated validation summary: continuous error, reliability, latency, token, and cost metrics.}
\label{tab:validation-120-ops}
\label{tab:cost-reliability}\\
\toprule
\textbf{System} & \textbf{MAE} & \textbf{Valid JSON} & \textbf{Failure} & \textbf{Avg. latency} & \textbf{Avg. tokens} & \textbf{Cost/case} & \textbf{Total cost} \\
\midrule
\endfirsthead
\toprule
\textbf{System} & \textbf{MAE} & \textbf{Valid JSON} & \textbf{Failure} & \textbf{Avg. latency} & \textbf{Avg. tokens} & \textbf{Cost/case} & \textbf{Total cost} \\
\midrule
\endhead
\bottomrule
\endlastfoot
Nanobot & 56.617 & 1.000 & 0.000 & 34.287 & 7968.908 & 0.003 & 0.345 \\
LangGraph & 58.683 & 1.000 & 0.000 & 72.562 & 27529.942 & 0.010 & 1.212 \\
NanoClaw & 66.358 & 1.000 & 0.000 & 29.351 & 14029.575 & 0.005 & 0.627 \\
Hermes AI & 57.858 & 1.000 & 0.000 & 33.198 & 8031.433 & 0.003 & 0.348 \\
OpenClaw & 62.000 & 1.000 & 0.000 & 30.742 & 13989.475 & 0.005 & 0.625 \\
EvoAgentX & 66.917 & 1.000 & 0.000 & 54.254 & 14096.733 & 0.005 & 0.611 \\
Historical persistence & 148.858 & 1.000 & 0.000 & 0.000 & 0.000 & 0.000 & 0.000 \\
Threshold rule & 148.890 & 1.000 & 0.000 & 0.000 & 0.000 & 0.000 & 0.000 \\
Always stable & 63.908 & 1.000 & 0.000 & 0.000 & 0.000 & 0.000 & 0.000 \\
Simple ML logistic & 113.500 & 1.000 & 0.000 & 0.000 & 0.000 & 0.000 & 0.000 \\
\end{longtable}
}

\newpage
\subsection{120-Case Stress and Ablation Diagnostics}
\label{app:calibration-verifier-anonymization}
\label{app:ablations}

The calibrated-label variant applies a frozen rule defined before the 120-case intermediate validation. It changes only the mapped class label from the model's own predicted depeg probability and predicted maximum absolute deviation; it does not modify raw model outputs, validation labels, future prices, or hidden horizon outcomes. The verifier ablation adds one risk-auditor call that receives the same lookback packet and first prediction, and may revise labels or probabilities only from lookback-window evidence.

{\StableEvalRoomyTable
\begin{longtable}{@{}p{0.34\linewidth}p{0.58\linewidth}@{}}
\caption{Frozen calibrated-label and verifier rules used for the 120-case ablation and reused for post-hoc calibration in the 507-case block.}
\label{tab:calibration-verifier-rules}\\
\toprule
\textbf{Component} & \textbf{Rule} \\
\midrule
\endfirsthead
\toprule
\textbf{Component} & \textbf{Rule} \\
\midrule
\endhead
\bottomrule
\endlastfoot
Stress rule 1 & Predict \texttt{stress} if \texttt{predicted\_max\_abs\_deviation\_bps} $\ge 100$. \\
Stress rule 2 & Predict \texttt{stress} if \texttt{p\_depeg\_next\_7d} $\ge 0.35$ and \texttt{predicted\_max\_abs\_deviation\_bps} $\ge 75$. \\
Watch rule & Predict \texttt{watch} if \texttt{p\_depeg\_next\_7d} $\ge 0.15$ or \texttt{predicted\_max\_abs\_deviation\_bps} $\ge 40$. \\
Stable rule & Predict \texttt{stable} otherwise. \\
Verifier input & Same 30-day lookback packet plus the first prediction. \\
Verifier constraint & May revise only from lookback-window evidence; no future data, hidden labels, or web browsing. \\
\end{longtable}
}

{\StableEvalCompactTable
\begin{longtable}{@{}p{0.20\linewidth}p{0.10\linewidth}rrrrr@{}}
\caption{120-case stress-class recall plus strict sustained-depeg TP/FN audit. Binary TP/FN refer only to the three strict sustained-depeg cases, while stress recall is the three-class stress-label recall. Stress R is recall on the three-class stress label. MDR and Binary TP/FN are computed only over the three strict sustained-depeg cases; FAR is computed over non-strict-depeg cases.}
\label{tab:stress-recall}\\
\toprule
\textbf{System} & \textbf{Type} & \textbf{Stress R} & \textbf{MDR} & \textbf{FAR} & \textbf{Binary TP} & \textbf{Binary FN} \\
\midrule
\endfirsthead
\toprule
\textbf{System} & \textbf{Type} & \textbf{Stress R} & \textbf{MDR} & \textbf{FAR} & \textbf{Binary TP} & \textbf{Binary FN} \\
\midrule
\endhead
\bottomrule
\endlastfoot
Historical persistence & baseline & 0.333 & 0.667 & 0.376 & 1 & 2 \\
Threshold rule & baseline & 0.333 & 1.000 & 0.282 & 0 & 3 \\
Nanobot & agent & 0.095 & 1.000 & 0.009 & 0 & 3 \\
OpenClaw & agent & 0.095 & 1.000 & 0.017 & 0 & 3 \\
LangGraph & agent & 0.048 & 1.000 & 0.009 & 0 & 3 \\
NanoClaw & agent & 0.048 & 1.000 & 0.017 & 0 & 3 \\
Hermes AI & agent & 0.048 & 1.000 & 0.017 & 0 & 3 \\
EvoAgentX & agent & 0.048 & 1.000 & 0.017 & 0 & 3 \\
Always stable & baseline & 0.000 & 1.000 & 0.000 & 0 & 3 \\
Simple ML logistic & baseline & 0.000 & 1.000 & 0.000 & 0 & 3 \\
\end{longtable}
}

{\StableEvalRoomyTable
\begin{longtable}{@{}p{0.28\linewidth}rrrr@{}}
\caption{120-case validation ablation summary. Calibration improves the best aggregate metrics without additional calls; the verifier improves best stress recall but raises cost.}
\label{tab:ablation-summary}\\
\toprule
\textbf{Variant} & \textbf{Best Macro-F1} & \textbf{Best BalAcc} & \textbf{Best Stress R} & \textbf{Mean cost} \\
\midrule
\endfirsthead
\toprule
\textbf{Variant} & \textbf{Best Macro-F1} & \textbf{Best BalAcc} & \textbf{Best Stress R} & \textbf{Mean cost} \\
\midrule
\endhead
\bottomrule
\endlastfoot
base-calibrated & 0.351 & 0.393 & 0.095 & 0.628 \\
base-raw & 0.301 & 0.368 & 0.000 & 0.628 \\
verifier-calibrated & 0.349 & 0.388 & 0.143 & 1.260 \\
verifier-raw & 0.272 & 0.350 & 0.095 & 1.260 \\
\end{longtable}
}

\StableEvalLandscapeStart
\begin{longtable}{@{}p{0.15\linewidth}p{0.17\linewidth}p{0.08\linewidth}rrrrrr@{}}
\caption{120-case full leaderboard across validation variants: classification metrics.}
\label{tab:appendix-leaderboard}\\
\toprule
\textbf{Variant} & \textbf{System} & \textbf{Type} & \textbf{Acc.} & \textbf{Macro-F1} & \textbf{BalAcc} & \textbf{R stable} & \textbf{R watch} & \textbf{R stress} \\
\midrule
\endfirsthead
\toprule
\textbf{Variant} & \textbf{System} & \textbf{Type} & \textbf{Acc.} & \textbf{Macro-F1} & \textbf{BalAcc} & \textbf{R stable} & \textbf{R watch} & \textbf{R stress} \\
\midrule
\endhead
\bottomrule
\endlastfoot
base-calibrated & Nanobot & agent & 0.442 & 0.351 & 0.393 & 0.241 & 0.844 & 0.095 \\
base-calibrated & LangGraph & agent & 0.442 & 0.337 & 0.381 & 0.296 & 0.800 & 0.048 \\
base-calibrated & NanoClaw & agent & 0.442 & 0.328 & 0.385 & 0.241 & 0.867 & 0.048 \\
base-calibrated & Hermes AI & agent & 0.425 & 0.321 & 0.369 & 0.259 & 0.800 & 0.048 \\
base-calibrated & OpenClaw & agent & 0.408 & 0.318 & 0.368 & 0.185 & 0.822 & 0.095 \\
base-calibrated & Historical persistence & baseline & 0.292 & 0.285 & 0.300 & 0.278 & 0.289 & 0.333 \\
base-calibrated & Threshold rule & baseline & 0.283 & 0.278 & 0.295 & 0.241 & 0.311 & 0.333 \\
base-calibrated & EvoAgentX & agent & 0.392 & 0.272 & 0.348 & 0.130 & 0.867 & 0.048 \\
base-calibrated & Always stable & baseline & 0.450 & 0.207 & 0.333 & 1.000 & 0.000 & 0.000 \\
base-calibrated & Simple ML logistic & baseline & 0.450 & 0.207 & 0.333 & 1.000 & 0.000 & 0.000 \\
base-raw & LangGraph & agent & 0.433 & 0.301 & 0.368 & 0.259 & 0.844 & 0.000 \\
base-raw & NanoClaw & agent & 0.425 & 0.286 & 0.364 & 0.204 & 0.889 & 0.000 \\
base-raw & Historical persistence & baseline & 0.292 & 0.285 & 0.300 & 0.278 & 0.289 & 0.333 \\
base-raw & Hermes AI & agent & 0.425 & 0.281 & 0.365 & 0.185 & 0.911 & 0.000 \\
base-raw & OpenClaw & agent & 0.417 & 0.281 & 0.357 & 0.204 & 0.867 & 0.000 \\
base-raw & Threshold rule & baseline & 0.283 & 0.278 & 0.295 & 0.241 & 0.311 & 0.333 \\
base-raw & Nanobot & agent & 0.400 & 0.266 & 0.343 & 0.185 & 0.844 & 0.000 \\
base-raw & EvoAgentX & agent & 0.383 & 0.237 & 0.333 & 0.111 & 0.889 & 0.000 \\
base-raw & Always stable & baseline & 0.450 & 0.207 & 0.333 & 1.000 & 0.000 & 0.000 \\
base-raw & Simple ML logistic & baseline & 0.450 & 0.207 & 0.333 & 1.000 & 0.000 & 0.000 \\
baselines-only & Historical persistence & baseline & 0.292 & 0.285 & 0.300 & 0.278 & 0.289 & 0.333 \\
baselines-only & Threshold rule & baseline & 0.283 & 0.278 & 0.295 & 0.241 & 0.311 & 0.333 \\
baselines-only & Always stable & baseline & 0.450 & 0.207 & 0.333 & 1.000 & 0.000 & 0.000 \\
baselines-only & Simple ML logistic & baseline & 0.450 & 0.207 & 0.333 & 1.000 & 0.000 & 0.000 \\
verifier-calibrated & Nanobot & agent & 0.425 & 0.349 & 0.388 & 0.222 & 0.800 & 0.143 \\
verifier-calibrated & Hermes AI & agent & 0.417 & 0.348 & 0.380 & 0.241 & 0.756 & 0.143 \\
verifier-calibrated & NanoClaw & agent & 0.400 & 0.315 & 0.359 & 0.204 & 0.778 & 0.095 \\
verifier-calibrated & LangGraph & agent & 0.383 & 0.312 & 0.342 & 0.241 & 0.689 & 0.095 \\
verifier-calibrated & Historical persistence & baseline & 0.292 & 0.285 & 0.300 & 0.278 & 0.289 & 0.333 \\
verifier-calibrated & Threshold rule & baseline & 0.283 & 0.278 & 0.295 & 0.241 & 0.311 & 0.333 \\
verifier-calibrated & OpenClaw & agent & 0.358 & 0.277 & 0.326 & 0.148 & 0.733 & 0.095 \\
verifier-calibrated & EvoAgentX & agent & 0.358 & 0.266 & 0.328 & 0.111 & 0.778 & 0.095 \\
verifier-calibrated & Always stable & baseline & 0.450 & 0.207 & 0.333 & 1.000 & 0.000 & 0.000 \\
verifier-calibrated & Simple ML logistic & baseline & 0.450 & 0.207 & 0.333 & 1.000 & 0.000 & 0.000 \\
verifier-raw & Historical persistence & baseline & 0.292 & 0.285 & 0.300 & 0.278 & 0.289 & 0.333 \\
verifier-raw & Threshold rule & baseline & 0.283 & 0.278 & 0.295 & 0.241 & 0.311 & 0.333 \\
verifier-raw & LangGraph & agent & 0.383 & 0.272 & 0.341 & 0.130 & 0.844 & 0.048 \\
verifier-raw & NanoClaw & agent & 0.367 & 0.271 & 0.335 & 0.111 & 0.800 & 0.095 \\
verifier-raw & EvoAgentX & agent & 0.375 & 0.262 & 0.345 & 0.074 & 0.867 & 0.095 \\
verifier-raw & Hermes AI & agent & 0.392 & 0.261 & 0.350 & 0.093 & 0.911 & 0.048 \\
verifier-raw & Nanobot & agent & 0.375 & 0.253 & 0.336 & 0.093 & 0.867 & 0.048 \\
verifier-raw & OpenClaw & agent & 0.367 & 0.241 & 0.319 & 0.111 & 0.844 & 0.000 \\
verifier-raw & Always stable & baseline & 0.450 & 0.207 & 0.333 & 1.000 & 0.000 & 0.000 \\
verifier-raw & Simple ML logistic & baseline & 0.450 & 0.207 & 0.333 & 1.000 & 0.000 & 0.000 \\
\end{longtable}
\StableEvalLandscapeEnd

\StableEvalLandscapeStart
\begin{longtable}{@{}p{0.15\linewidth}p{0.17\linewidth}rrrrrrr@{}}
\caption{120-case full leaderboard across validation variants: risk, continuous-error, and cost metrics.}
\label{tab:appendix-leaderboard-risk-cost}\\
\toprule
\textbf{Variant} & \textbf{System} & \textbf{MDR} & \textbf{FAR} & \textbf{Brier} & \textbf{MAE} & \textbf{RMSE} & \textbf{Cost} & \textbf{Cost/correct} \\
\midrule
\endfirsthead
\toprule
\textbf{Variant} & \textbf{System} & \textbf{MDR} & \textbf{FAR} & \textbf{Brier} & \textbf{MAE} & \textbf{RMSE} & \textbf{Cost} & \textbf{Cost/correct} \\
\midrule
\endhead
\bottomrule
\endlastfoot
base-calibrated & Nanobot & 1.000 & 0.009 & 0.047 & 56.617 & 167.172 & 0.345 & 0.007 \\
base-calibrated & LangGraph & 1.000 & 0.009 & 0.046 & 58.683 & 168.627 & 1.212 & 0.023 \\
base-calibrated & NanoClaw & 1.000 & 0.017 & 0.050 & 66.358 & 199.767 & 0.627 & 0.012 \\
base-calibrated & Hermes AI & 1.000 & 0.017 & 0.047 & 57.858 & 174.954 & 0.348 & 0.007 \\
base-calibrated & OpenClaw & 1.000 & 0.017 & 0.048 & 62.000 & 180.325 & 0.625 & 0.013 \\
base-calibrated & Historical persistence & 0.667 & 0.376 & 0.276 & 148.858 & 309.496 & 0.000 & 0.000 \\
base-calibrated & Threshold rule & 1.000 & 0.282 & 0.244 & 148.890 & 309.496 & 0.000 & 0.000 \\
base-calibrated & EvoAgentX & 1.000 & 0.017 & 0.050 & 66.917 & 199.181 & 0.611 & 0.013 \\
base-calibrated & Always stable & 1.000 & 0.000 & 0.025 & 63.908 & 148.390 & 0.000 & 0.000 \\
base-calibrated & Simple ML logistic & 1.000 & 0.000 & 0.024 & 113.500 & 368.523 & 0.000 & 0.000 \\
base-raw & LangGraph & 1.000 & 0.009 & 0.046 & 58.683 & 168.627 & 1.212 & 0.023 \\
base-raw & NanoClaw & 1.000 & 0.017 & 0.050 & 66.358 & 199.767 & 0.627 & 0.012 \\
base-raw & Historical persistence & 0.667 & 0.376 & 0.276 & 148.858 & 309.496 & 0.000 & 0.000 \\
base-raw & Hermes AI & 1.000 & 0.017 & 0.047 & 57.858 & 174.954 & 0.348 & 0.007 \\
base-raw & OpenClaw & 1.000 & 0.017 & 0.048 & 62.000 & 180.325 & 0.625 & 0.012 \\
base-raw & Threshold rule & 1.000 & 0.282 & 0.244 & 148.890 & 309.496 & 0.000 & 0.000 \\
base-raw & Nanobot & 1.000 & 0.009 & 0.047 & 56.617 & 167.172 & 0.345 & 0.007 \\
base-raw & EvoAgentX & 1.000 & 0.017 & 0.050 & 66.917 & 199.181 & 0.611 & 0.013 \\
base-raw & Always stable & 1.000 & 0.000 & 0.025 & 63.908 & 148.390 & 0.000 & 0.000 \\
base-raw & Simple ML logistic & 1.000 & 0.000 & 0.024 & 113.500 & 368.523 & 0.000 & 0.000 \\
baselines-only & Historical persistence & 0.667 & 0.376 & 0.276 & 148.858 & 309.496 & 0.000 & 0.000 \\
baselines-only & Threshold rule & 1.000 & 0.282 & 0.244 & 148.890 & 309.496 & 0.000 & 0.000 \\
baselines-only & Always stable & 1.000 & 0.000 & 0.025 & 63.908 & 148.390 & 0.000 & 0.000 \\
baselines-only & Simple ML logistic & 1.000 & 0.000 & 0.024 & 113.500 & 368.523 & 0.000 & 0.000 \\
verifier-calibrated & Nanobot & 1.000 & 0.017 & 0.052 & 59.992 & 168.163 & 0.978 & 0.019 \\
verifier-calibrated & Hermes AI & 1.000 & 0.017 & 0.051 & 60.608 & 176.005 & 0.982 & 0.020 \\
verifier-calibrated & NanoClaw & 1.000 & 0.017 & 0.054 & 71.158 & 200.912 & 1.257 & 0.026 \\
verifier-calibrated & LangGraph & 1.000 & 0.017 & 0.054 & 65.442 & 185.971 & 1.840 & 0.040 \\
verifier-calibrated & Historical persistence & 0.667 & 0.376 & 0.276 & 148.858 & 309.496 & 0.000 & 0.000 \\
verifier-calibrated & Threshold rule & 1.000 & 0.282 & 0.244 & 148.890 & 309.496 & 0.000 & 0.000 \\
verifier-calibrated & OpenClaw & 1.000 & 0.017 & 0.053 & 70.850 & 200.598 & 1.253 & 0.029 \\
verifier-calibrated & EvoAgentX & 1.000 & 0.017 & 0.053 & 69.217 & 199.638 & 1.246 & 0.029 \\
verifier-calibrated & Always stable & 1.000 & 0.000 & 0.025 & 63.908 & 148.390 & 0.000 & 0.000 \\
verifier-calibrated & Simple ML logistic & 1.000 & 0.000 & 0.024 & 113.500 & 368.523 & 0.000 & 0.000 \\
verifier-raw & Historical persistence & 0.667 & 0.376 & 0.276 & 148.858 & 309.496 & 0.000 & 0.000 \\
verifier-raw & Threshold rule & 1.000 & 0.282 & 0.244 & 148.890 & 309.496 & 0.000 & 0.000 \\
verifier-raw & LangGraph & 1.000 & 0.017 & 0.054 & 65.442 & 185.971 & 1.840 & 0.040 \\
verifier-raw & NanoClaw & 1.000 & 0.017 & 0.054 & 71.158 & 200.912 & 1.257 & 0.029 \\
verifier-raw & EvoAgentX & 1.000 & 0.017 & 0.053 & 69.217 & 199.638 & 1.246 & 0.028 \\
verifier-raw & Hermes AI & 1.000 & 0.017 & 0.051 & 60.608 & 176.005 & 0.982 & 0.021 \\
verifier-raw & Nanobot & 1.000 & 0.017 & 0.052 & 59.992 & 168.163 & 0.978 & 0.022 \\
verifier-raw & OpenClaw & 1.000 & 0.017 & 0.053 & 70.850 & 200.598 & 1.253 & 0.028 \\
verifier-raw & Always stable & 1.000 & 0.000 & 0.025 & 63.908 & 148.390 & 0.000 & 0.000 \\
verifier-raw & Simple ML logistic & 1.000 & 0.000 & 0.024 & 113.500 & 368.523 & 0.000 & 0.000 \\
\end{longtable}
\StableEvalLandscapeEnd

\section{StableEval-507-Natural Scaling Results}
\label{app:scaling-507-results}
\label{app:stableeval-507}

The 507-case natural-distribution block contains 441 stable cases, 45 watch cases, and 21 stress cases. It uses the same stablecoin universe, 30-day lookback window, hidden 7-day horizon, risk-sensitive prompt contract, OpenRouter Qwen backend, temperature-zero decoding, strict JSON parsing, and frozen post-hoc calibration rule used in the 120-case block. The stable-heavy composition makes this block an operational scaling and robustness check rather than a replacement for the stress-enriched 120-case validation.

\subsection{507-Case Leaderboard-Style Figures}
\label{app:scaling-507-figures}

The following figures mirror the 120-case summaries: calibrated leaderboard, raw-versus-calibrated ablation, and cost-aware performance tradeoffs. Because the full 507-case pool is stable-heavy, non-agent baselines are comparatively strong on aggregate Macro-F1; this is an expected distribution effect.

\begin{figure}[!htbp]
\centering
\begin{tikzpicture}
\begin{axis}[
    xbar,
    width=0.90\linewidth,
    height=0.38\linewidth,
    xmin=0,
    xmax=0.54,
    bar width=2.8pt,
    enlarge y limits=0.065,
    symbolic y coords={
        Simple ML logistic,
        Historical persistence,
        Threshold rule,
        Always stable,
        LangGraph,
        NanoClaw,
        Hermes AI,
        Nanobot,
        EvoAgentX,
        OpenClaw
    },
    ytick=data,
    yticklabel style={font=\scriptsize, text width=2.9cm, align=right},
    xtick={0,0.1,0.2,0.3,0.4,0.5},
    xlabel={Score},
    xlabel style={font=\small},
    tick label style={font=\scriptsize},
    xmajorgrids=true,
    grid style={draw=stableGray!22},
    axis line style={draw=stableGray!45},
    legend style={
        draw=stableGray!35,
        fill=white,
        font=\scriptsize,
        at={(0.5,1.04)},
        anchor=south,
        legend columns=3
    },
    extra x ticks={0.4019451812555261},
    extra x tick labels={Simple ML logistic},
    extra x tick style={
        tick label style={font=\scriptsize, rotate=90},
        grid=major,
        major grid style={draw=stableNavy, densely dashed, line width=0.8pt}
    },
]
\addplot+[draw=stableData!80!black, fill=stableData!45] coordinates {
    (0.4019451812555261,Simple ML logistic)
    (0.3866953108284538,Historical persistence)
    (0.3763071106821107,Threshold rule)
    (0.310126582278481,Always stable)
    (0.304429361488185,LangGraph)
    (0.3041209184684067,NanoClaw)
    (0.301050078943466,Hermes AI)
    (0.2873804156351195,Nanobot)
    (0.2625568759541383,EvoAgentX)
    (0.2484606677985285,OpenClaw)
};
\addplot+[draw=stableProtocol!80!black, fill=stableProtocol!45] coordinates {
    (0.4929705215419501,Simple ML logistic)
    (0.4500377928949357,Historical persistence)
    (0.4461073318216175,Threshold rule)
    (0.3333333333333333,Always stable)
    (0.4403628117913832,LangGraph)
    (0.4352229780801209,NanoClaw)
    (0.4328042328042327,Hermes AI)
    (0.4275132275132274,Nanobot)
    (0.4137566137566137,EvoAgentX)
    (0.4031746031746031,OpenClaw)
};
\addplot+[draw=stableRisk!80!black, fill=stableRisk!45] coordinates {
    (0.3809523809523809,Simple ML logistic)
    (0.3333333333333333,Historical persistence)
    (0.3333333333333333,Threshold rule)
    (0,Always stable)
    (0.0476190476190476,LangGraph)
    (0.04761904761904762,NanoClaw)
    (0.09523809523809521,Hermes AI)
    (0.09523809523809521,Nanobot)
    (0.0476190476190476,EvoAgentX)
    (0.0476190476190476,OpenClaw)
};
\legend{Macro-F1, Balanced accuracy, Stress recall}
\end{axis}
\end{tikzpicture}
\caption{Full 507-case natural-distribution leaderboard. Bars report Macro-F1, balanced accuracy, and stress-class recall from the calibrated first-pass protocol. The dashed reference line marks the Macro-F1 of the strongest non-agent baseline, Simple ML logistic.}
\label{fig:scaling-507-leaderboard}
\end{figure}
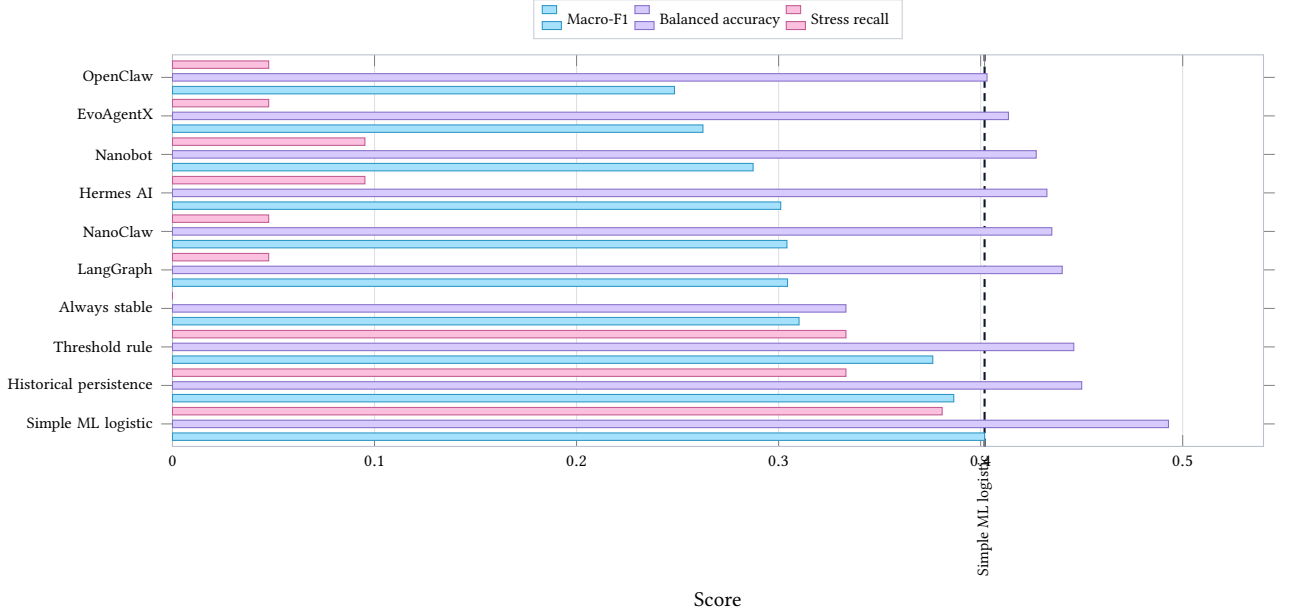

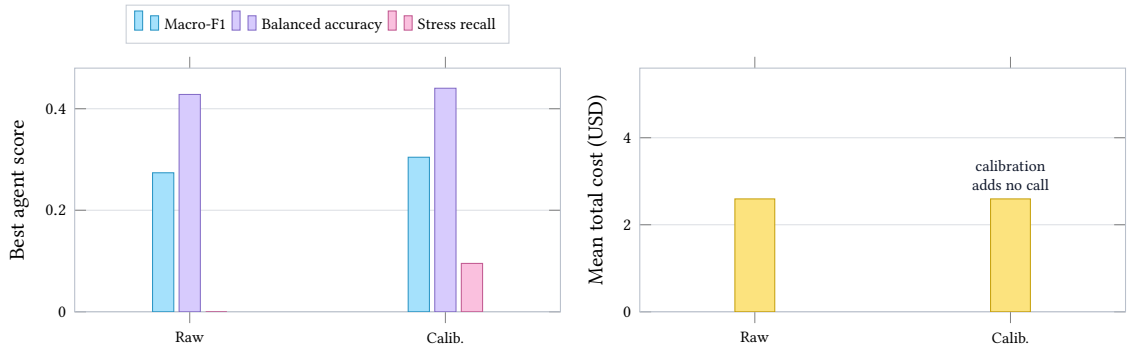
\begin{figure}[!htbp]
\centering
\begin{tikzpicture}
\begin{groupplot}[
    group style={group size=2 by 1, horizontal sep=1.05cm},
    width=0.45\linewidth,
    height=0.27\linewidth,
    tick label style={font=\scriptsize},
    label style={font=\small},
    axis line style={draw=stableGray!45},
    ymajorgrids=true,
    grid style={draw=stableGray!22},
]
\nextgroupplot[
    ybar,
    ymin=0,
    ymax=0.48,
    bar width=8pt,
    xmin=0.55,
    xmax=2.45,
    ylabel={Best agent score},
    xtick={1,2},
    xticklabels={Raw, Calib.},
    legend style={
        draw=stableGray!35,
        fill=white,
        font=\scriptsize,
        at={(0.5,1.10)},
        anchor=south,
        legend columns=3
    },
]
\addplot+[draw=stableData!80!black, fill=stableData!45] coordinates {
    (1,0.2738348816534507)
    (2,0.304429361488185)
};
\addplot+[draw=stableProtocol!80!black, fill=stableProtocol!45] coordinates {
    (1,0.4281179138321995)
    (2,0.4403628117913832)
};
\addplot+[draw=stableRisk!80!black, fill=stableRisk!45] coordinates {
    (1,0)
    (2,0.09523809523809521)
};
\legend{Macro-F1, Balanced accuracy, Stress recall}

\nextgroupplot[
    ybar,
    ymin=0,
    ymax=5.6,
    bar width=15pt,
    xmin=0.55,
    xmax=2.45,
    ylabel={Mean total cost (USD)},
    xtick={1,2},
    xticklabels={Raw, Calib.},
]
\addplot+[draw=stableAmber!80!black, fill=stableAmber!55] coordinates {
    (1,2.5938891)
    (2,2.5938891)
};
\node[font=\scriptsize, align=center, text=stableNavy] at (axis cs:2,3.1438891) {calibration\\adds no call};
\end{groupplot}
\end{tikzpicture}
\caption{507-case raw-versus-calibrated ablation. Calibration improves the best agent Macro-F1, balanced accuracy, and stress recall without increasing total cost because it remaps already returned risk estimates and predicted deviations.}
\label{fig:scaling-507-ablation}
\end{figure}

\begin{figure}[!htbp]
\centering
\begin{tikzpicture}
\begin{groupplot}[
    group style={group size=2 by 1, horizontal sep=1.05cm},
    width=0.45\linewidth,
    height=0.27\linewidth,
    ymin=0.22,
    ymax=0.43,
    tick label style={font=\scriptsize},
    label style={font=\small},
    axis line style={draw=stableGray!45},
    xmajorgrids=true,
    ymajorgrids=true,
    grid style={draw=stableGray!22},
]
\nextgroupplot[
    xmin=1.2,
    xmax=5.25,
    xlabel={Total cost (USD)},
    ylabel={Macro-F1},
]
\addplot[draw=stableNavy, densely dashed, line width=0.8pt, forget plot] coordinates {
    (1.2,0.4019451812555261)
    (5.25,0.4019451812555261)
};
\node[font=\scriptsize, anchor=south west, text=stableNavy] at (axis cs:3.25,0.4019451812555261) {best baseline};
\addplot+[only marks, mark=*, mark size=2.5pt, draw=stableNavy, fill=stableAmber] coordinates {(1.47100392,0.2873804156351195)};
\addplot+[only marks, mark=*, mark size=2.5pt, draw=stableNavy, fill=stableGray] coordinates {(4.98023764,0.304429361488185)};
\addplot+[only marks, mark=*, mark size=2.5pt, draw=stableNavy, fill=stableEval] coordinates {(2.54122352,0.3041209184684067)};
\addplot+[only marks, mark=*, mark size=2.5pt, draw=stableNavy, fill=stableEvalCard] coordinates {(1.47384064,0.301050078943466)};
\addplot+[only marks, mark=*, mark size=2.5pt, draw=stableNavy, fill=stableData] coordinates {(2.54528804,0.2484606677985285)};
\addplot+[only marks, mark=*, mark size=2.5pt, draw=stableNavy, fill=stableProtocol] coordinates {(2.55174084,0.2625568759541383)};
\node[font=\scriptsize, anchor=west] at (axis cs:1.55100392,0.2873804156351195) {Nanobot};
\node[font=\scriptsize, anchor=west] at (axis cs:4.430237640000001,0.312429361488185) {LangGraph};
\node[font=\scriptsize, anchor=west] at (axis cs:2.60122352,0.2961209184684067) {NanoClaw};
\node[font=\scriptsize, anchor=west] at (axis cs:1.55384064,0.307050078943466) {Hermes};
\node[font=\scriptsize, anchor=west] at (axis cs:2.61528804,0.2424606677985285) {OpenClaw};
\node[font=\scriptsize, anchor=west] at (axis cs:2.63174084,0.2625568759541383) {EvoAgentX};

\nextgroupplot[
    xmin=19,
    xmax=37,
    xlabel={Average latency (seconds)},
    yticklabels={},
]
\addplot[draw=stableNavy, densely dashed, line width=0.8pt, forget plot] coordinates {
    (19,0.4019451812555261)
    (37,0.4019451812555261)
};
\node[font=\scriptsize, anchor=south west, text=stableNavy] at (axis cs:29.5,0.4019451812555261) {best baseline};
\addplot+[only marks, mark=square*, mark size=2.5pt, draw=stableNavy, fill=stableAmber] coordinates {(28.43145539727058,0.2873804156351195)};
\addplot+[only marks, mark=square*, mark size=2.5pt, draw=stableNavy, fill=stableGray] coordinates {(34.83513575684542,0.304429361488185)};
\addplot+[only marks, mark=square*, mark size=2.5pt, draw=stableNavy, fill=stableEval] coordinates {(21.0566847044889,0.3041209184684067)};
\addplot+[only marks, mark=square*, mark size=2.5pt, draw=stableNavy, fill=stableEvalCard] coordinates {(25.59679733983393,0.301050078943466)};
\addplot+[only marks, mark=square*, mark size=2.5pt, draw=stableNavy, fill=stableData] coordinates {(23.59686400684641,0.2484606677985285)};
\addplot+[only marks, mark=square*, mark size=2.5pt, draw=stableNavy, fill=stableProtocol] coordinates {(27.71971019145439,0.2625568759541383)};
\node[font=\scriptsize, anchor=west] at (axis cs:28.93145539727058,0.2873804156351195) {Nanobot};
\node[font=\scriptsize, anchor=west] at (axis cs:31.63513575684543,0.312429361488185) {LangGraph};
\node[font=\scriptsize, anchor=west] at (axis cs:21.5566847044889,0.2961209184684067) {NanoClaw};
\node[font=\scriptsize, anchor=west] at (axis cs:26.09679733983393,0.307050078943466) {Hermes};
\node[font=\scriptsize, anchor=west] at (axis cs:24.09686400684641,0.2424606677985285) {OpenClaw};
\node[font=\scriptsize, anchor=west] at (axis cs:28.21971019145439,0.2625568759541383) {EvoAgentX};
\end{groupplot}
\end{tikzpicture}
\caption{Cost-aware performance tradeoffs for the six agentic systems on the full 507-case natural-distribution scaling check. The agents remain operationally reliable, but the stable-heavy distribution makes the strongest non-agent baseline difficult to beat on aggregate Macro-F1.}
\label{fig:scaling-507-cost-performance}
\end{figure}
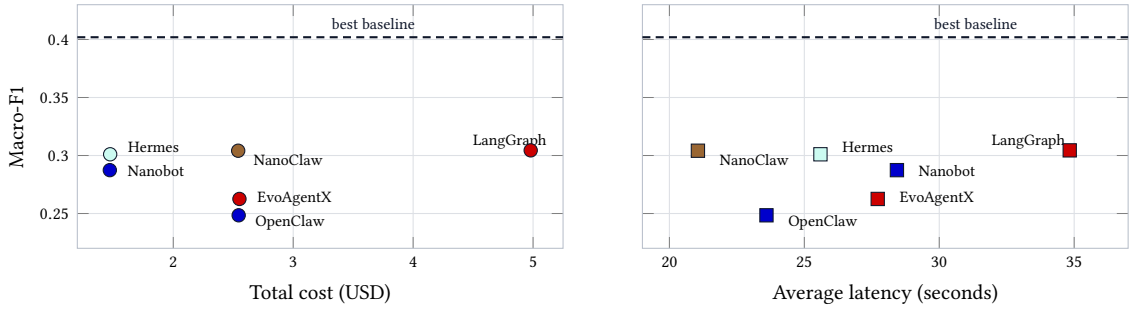

\newpage
\subsection{507-Case Leaderboard Tables}
\label{app:scaling-507-tables}

Tables~\ref{tab:scaling-507} and~\ref{tab:scaling-507-ops} report the calibrated 507-case summary. The raw/calibrated matrices remain in this section so reviewers can reconstruct the complete six-agent and four-baseline scaling result.

{\StableEvalCompactTable
\begin{longtable}{@{}p{0.18\linewidth}p{0.09\linewidth}rrrrrr@{}}
\caption{507-case natural-distribution scaling summary: predictive and risk metrics.}
\label{tab:scaling-507}
\label{tab:agent-baseline-507}\\
\toprule
\textbf{System} & \textbf{Type} & \textbf{Acc.} & \textbf{Macro-F1} & \textbf{BalAcc} & \textbf{Stress R} & \textbf{MDR} & \textbf{FAR} \\
\midrule
\endfirsthead
\toprule
\textbf{System} & \textbf{Type} & \textbf{Acc.} & \textbf{Macro-F1} & \textbf{BalAcc} & \textbf{Stress R} & \textbf{MDR} & \textbf{FAR} \\
\midrule
\endhead
\bottomrule
\endlastfoot
LangGraph & agent & 0.467 & 0.304 & 0.440 & 0.048 & 1.000 & 0.004 \\
NanoClaw & agent & 0.471 & 0.304 & 0.435 & 0.048 & 1.000 & 0.004 \\
Hermes AI & agent & 0.408 & 0.301 & 0.433 & 0.095 & 1.000 & 0.004 \\
Nanobot & agent & 0.394 & 0.287 & 0.428 & 0.095 & 1.000 & 0.002 \\
EvoAgentX & agent & 0.381 & 0.263 & 0.414 & 0.048 & 1.000 & 0.004 \\
OpenClaw & agent & 0.353 & 0.248 & 0.403 & 0.048 & 1.000 & 0.004 \\
Simple ML logistic & baseline & 0.641 & 0.402 & 0.493 & 0.381 & 1.000 & 0.077 \\
Historical persistence & baseline & 0.673 & 0.387 & 0.450 & 0.333 & 0.667 & 0.149 \\
Threshold rule & baseline & 0.645 & 0.376 & 0.446 & 0.333 & 1.000 & 0.095 \\
Always stable & baseline & 0.870 & 0.310 & 0.333 & 0.000 & 1.000 & 0.000 \\
\end{longtable}
}

{\StableEvalCompactTable
\begin{longtable}{@{}p{0.17\linewidth}rrrrrrr@{}}
\caption{507-case natural-distribution scaling summary: continuous error, reliability, latency, token, and cost metrics.}
\label{tab:scaling-507-ops}
\label{tab:cost-reliability-507}\\
\toprule
\textbf{System} & \textbf{MAE} & \textbf{Valid JSON} & \textbf{Failure} & \textbf{Avg. latency} & \textbf{Avg. tokens} & \textbf{Cost/case} & \textbf{Total cost} \\
\midrule
\endfirsthead
\toprule
\textbf{System} & \textbf{MAE} & \textbf{Valid JSON} & \textbf{Failure} & \textbf{Avg. latency} & \textbf{Avg. tokens} & \textbf{Cost/case} & \textbf{Total cost} \\
\midrule
\endhead
\bottomrule
\endlastfoot
LangGraph & 31.063 & 1.000 & 0.000 & 34.835 & 27210.148 & 0.010 & 4.980 \\
NanoClaw & 31.672 & 1.000 & 0.000 & 21.057 & 13879.876 & 0.005 & 2.541 \\
Hermes AI & 34.375 & 1.000 & 0.000 & 25.597 & 8046.953 & 0.003 & 1.474 \\
Nanobot & 30.242 & 1.000 & 0.000 & 28.431 & 8031.456 & 0.003 & 1.471 \\
EvoAgentX & 32.680 & 1.000 & 0.000 & 27.720 & 13931.836 & 0.005 & 2.552 \\
OpenClaw & 34.655 & 1.000 & 0.000 & 23.597 & 13900.018 & 0.005 & 2.545 \\
Simple ML logistic & 17.067 & 1.000 & 0.000 & 0.000 & 0.000 & 0.000 & 0.000 \\
Historical persistence & 52.665 & 1.000 & 0.000 & 0.000 & 0.000 & 0.000 & 0.000 \\
Threshold rule & 52.688 & 1.000 & 0.000 & 0.000 & 0.000 & 0.000 & 0.000 \\
Always stable & 20.170 & 1.000 & 0.000 & 0.000 & 0.000 & 0.000 & 0.000 \\
\end{longtable}
}

\newpage
\subsection{507-Case Stress and Calibration Diagnostics}
\label{app:scaling-507-calibration-context}

The 507-case scaling run separates raw model labels from calibrated labels in the same way as the 120-case block. Calibration uses the frozen rule in Table~\ref{tab:calibration-verifier-rules} and does not alter model probabilities, predicted maximum deviations, costs, tokens, or latencies. On the full natural-distribution set, calibration improves every agent's Macro-F1 relative to its raw first-pass label, while non-agent baselines remain strong because the full set is stable-heavy.

{\StableEvalCompactTable
\begin{longtable}{@{}p{0.20\linewidth}p{0.10\linewidth}rrrrr@{}}
\caption{507-case stress-class recall plus strict sustained-depeg TP/FN audit. Binary TP/FN refer only to the three strict sustained-depeg cases. Stress R is recall on the three-class stress label. MDR and Binary TP/FN are computed only over the three strict sustained-depeg cases; FAR is computed over non-strict-depeg cases.}
\label{tab:stress-recall-507}\\
\toprule
\textbf{System} & \textbf{Type} & \textbf{Stress R} & \textbf{MDR} & \textbf{FAR} & \textbf{Binary TP} & \textbf{Binary FN} \\
\midrule
\endfirsthead
\toprule
\textbf{System} & \textbf{Type} & \textbf{Stress R} & \textbf{MDR} & \textbf{FAR} & \textbf{Binary TP} & \textbf{Binary FN} \\
\midrule
\endhead
\bottomrule
\endlastfoot
Simple ML logistic & baseline & 0.381 & 1.000 & 0.077 & 0 & 3 \\
Historical persistence & baseline & 0.333 & 0.667 & 0.149 & 1 & 2 \\
Threshold rule & baseline & 0.333 & 1.000 & 0.095 & 0 & 3 \\
Hermes AI & agent & 0.095 & 1.000 & 0.004 & 0 & 3 \\
Nanobot & agent & 0.095 & 1.000 & 0.002 & 0 & 3 \\
NanoClaw & agent & 0.048 & 1.000 & 0.004 & 0 & 3 \\
LangGraph & agent & 0.048 & 1.000 & 0.004 & 0 & 3 \\
EvoAgentX & agent & 0.048 & 1.000 & 0.004 & 0 & 3 \\
OpenClaw & agent & 0.048 & 1.000 & 0.004 & 0 & 3 \\
Always stable & baseline & 0.000 & 1.000 & 0.000 & 0 & 3 \\
\end{longtable}
}

{\StableEvalRoomyTable
\begin{longtable}{@{}p{0.28\linewidth}rrrr@{}}
\caption{507-case raw-versus-calibrated ablation summary for agentic systems.}
\label{tab:ablation-summary-507}\\
\toprule
\textbf{Variant} & \textbf{Best agent Macro-F1} & \textbf{Best agent BalAcc} & \textbf{Best agent Stress R} & \textbf{Mean cost} \\
\midrule
\endfirsthead
\toprule
\textbf{Variant} & \textbf{Best agent Macro-F1} & \textbf{Best agent BalAcc} & \textbf{Best agent Stress R} & \textbf{Mean cost} \\
\midrule
\endhead
\bottomrule
\endlastfoot
full-507-calibrated & 0.304 & 0.440 & 0.095 & 2.594 \\
full-507-raw & 0.274 & 0.428 & 0.000 & 2.594 \\
\end{longtable}
}

{\StableEvalTinyTable
\begin{longtable}{@{}p{0.18\linewidth}p{0.19\linewidth}p{0.07\linewidth}rrrrrr@{}}
\caption{Full 507-case leaderboard: classification metrics for baselines and raw/calibrated agent variants.}
\label{tab:appendix-leaderboard-507}\\
\toprule
\textbf{Variant} & \textbf{System} & \textbf{Type} & \textbf{Acc.} & \textbf{Macro-F1} & \textbf{BalAcc} & \textbf{R stable} & \textbf{R watch} & \textbf{R stress} \\
\midrule
\endfirsthead
\toprule
\textbf{Variant} & \textbf{System} & \textbf{Type} & \textbf{Acc.} & \textbf{Macro-F1} & \textbf{BalAcc} & \textbf{R stable} & \textbf{R watch} & \textbf{R stress} \\
\midrule
\endhead
\bottomrule
\endlastfoot
full-507-calibrated & LangGraph & agent & 0.467 & 0.304 & 0.440 & 0.451 & 0.822 & 0.048 \\
full-507-calibrated & NanoClaw & agent & 0.471 & 0.304 & 0.435 & 0.458 & 0.800 & 0.048 \\
full-507-calibrated & Hermes AI & agent & 0.408 & 0.301 & 0.433 & 0.381 & 0.822 & 0.095 \\
full-507-calibrated & Nanobot & agent & 0.394 & 0.287 & 0.428 & 0.365 & 0.822 & 0.095 \\
full-507-calibrated & EvoAgentX & agent & 0.381 & 0.263 & 0.414 & 0.349 & 0.844 & 0.048 \\
full-507-calibrated & OpenClaw & agent & 0.353 & 0.248 & 0.403 & 0.317 & 0.844 & 0.048 \\
full-507-raw & LangGraph & agent & 0.458 & 0.274 & 0.428 & 0.440 & 0.844 & 0.000 \\
full-507-raw & NanoClaw & agent & 0.410 & 0.249 & 0.397 & 0.390 & 0.800 & 0.000 \\
full-507-raw & EvoAgentX & agent & 0.369 & 0.232 & 0.407 & 0.333 & 0.889 & 0.000 \\
full-507-raw & Nanobot & agent & 0.355 & 0.223 & 0.389 & 0.322 & 0.844 & 0.000 \\
full-507-raw & Hermes AI & agent & 0.335 & 0.214 & 0.388 & 0.297 & 0.867 & 0.000 \\
full-507-raw & OpenClaw & agent & 0.327 & 0.210 & 0.385 & 0.288 & 0.867 & 0.000 \\
full-507-baselines & Simple ML logistic & baseline & 0.641 & 0.402 & 0.493 & 0.676 & 0.422 & 0.381 \\
full-507-baselines & Historical persistence & baseline & 0.673 & 0.387 & 0.450 & 0.728 & 0.289 & 0.333 \\
full-507-baselines & Threshold rule & baseline & 0.645 & 0.376 & 0.446 & 0.694 & 0.311 & 0.333 \\
full-507-baselines & Always stable & baseline & 0.870 & 0.310 & 0.333 & 1.000 & 0.000 & 0.000 \\
\end{longtable}
}

{\StableEvalTinyTable
\begin{longtable}{@{}p{0.18\linewidth}p{0.20\linewidth}rrrrrrr@{}}
\caption{Full 507-case leaderboard: risk, continuous-error, and cost metrics for baselines and raw/calibrated agent variants.}
\label{tab:appendix-leaderboard-507-risk-cost}\\
\toprule
\textbf{Variant} & \textbf{System} & \textbf{MDR} & \textbf{FAR} & \textbf{Brier} & \textbf{MAE} & \textbf{RMSE} & \textbf{Cost} & \textbf{Cost/correct} \\
\midrule
\endfirsthead
\toprule
\textbf{Variant} & \textbf{System} & \textbf{MDR} & \textbf{FAR} & \textbf{Brier} & \textbf{MAE} & \textbf{RMSE} & \textbf{Cost} & \textbf{Cost/correct} \\
\midrule
\endhead
\bottomrule
\endlastfoot
full-507-calibrated & LangGraph & 1.000 & 0.004 & 0.021 & 31.063 & 99.662 & 4.980 & 0.021 \\
full-507-calibrated & NanoClaw & 1.000 & 0.004 & 0.020 & 31.672 & 100.153 & 2.541 & 0.011 \\
full-507-calibrated & Hermes AI & 1.000 & 0.004 & 0.019 & 34.375 & 99.966 & 1.474 & 0.007 \\
full-507-calibrated & Nanobot & 1.000 & 0.002 & 0.023 & 30.242 & 84.236 & 1.471 & 0.007 \\
full-507-calibrated & EvoAgentX & 1.000 & 0.004 & 0.022 & 32.680 & 100.002 & 2.552 & 0.013 \\
full-507-calibrated & OpenClaw & 1.000 & 0.004 & 0.023 & 34.655 & 100.163 & 2.545 & 0.014 \\
full-507-raw & LangGraph & 1.000 & 0.004 & 0.021 & 31.063 & 99.662 & 4.980 & 0.021 \\
full-507-raw & NanoClaw & 1.000 & 0.004 & 0.020 & 31.672 & 100.153 & 2.541 & 0.012 \\
full-507-raw & EvoAgentX & 1.000 & 0.004 & 0.022 & 32.680 & 100.002 & 2.552 & 0.014 \\
full-507-raw & Nanobot & 1.000 & 0.002 & 0.023 & 30.242 & 84.236 & 1.471 & 0.008 \\
full-507-raw & Hermes AI & 1.000 & 0.004 & 0.019 & 34.375 & 99.966 & 1.474 & 0.009 \\
full-507-raw & OpenClaw & 1.000 & 0.004 & 0.023 & 34.655 & 100.163 & 2.545 & 0.015 \\
full-507-baselines & Simple ML logistic & 1.000 & 0.077 & 0.067 & 17.067 & 72.335 & 0.000 & 0.000 \\
full-507-baselines & Historical persistence & 0.667 & 0.149 & 0.112 & 52.665 & 154.288 & 0.000 & 0.000 \\
full-507-baselines & Threshold rule & 1.000 & 0.095 & 0.093 & 52.688 & 154.287 & 0.000 & 0.000 \\
full-507-baselines & Always stable & 1.000 & 0.000 & 0.006 & 20.170 & 72.666 & 0.000 & 0.000 \\
\end{longtable}
}

\section{Cross-Block Robustness, Alignment, and Diagnostic Audits}
\label{app:cross-block-audits}

This section collects diagnostics that cut across result blocks or audit benchmark safety rather than reporting one block's main leaderboard. The purpose is to help reviewers distinguish aggregate performance, rare-stress behavior, calibration effects, leakage checks, and adapter caveats.

\subsection{120-Versus-507 Scaling Context}
\label{app:cross-block-alignment}

Table~\ref{tab:agent-120-507-context} compares the calibrated agentic systems across the stress-enriched and natural-distribution blocks. The 507 block has lower stress prevalence and many more stable controls, so changes in Macro-F1 and balanced accuracy should be interpreted as distribution-sensitive scaling behavior rather than a replacement for the 120-case validation.

{\StableEvalCompactTable
\begin{longtable}{@{}p{0.16\linewidth}rrrrrrrr@{}}
\caption{Compact 120-case versus 507-case calibrated scaling context for agentic systems.}
\label{tab:agent-120-507-context}\\
\toprule
\textbf{System} & \textbf{120 Macro-F1} & \textbf{507 Macro-F1} & \textbf{120 BalAcc} & \textbf{507 BalAcc} & \textbf{120 Stress R} & \textbf{507 Stress R} & \textbf{120 Cost} & \textbf{507 Cost} \\
\midrule
\endfirsthead
\toprule
\textbf{System} & \textbf{120 Macro-F1} & \textbf{507 Macro-F1} & \textbf{120 BalAcc} & \textbf{507 BalAcc} & \textbf{120 Stress R} & \textbf{507 Stress R} & \textbf{120 Cost} & \textbf{507 Cost} \\
\midrule
\endhead
\bottomrule
\endlastfoot
Nanobot & 0.351 & 0.287 & 0.393 & 0.428 & 0.095 & 0.095 & 0.345 & 1.471 \\
LangGraph & 0.337 & 0.304 & 0.381 & 0.440 & 0.048 & 0.048 & 1.212 & 4.980 \\
NanoClaw & 0.328 & 0.304 & 0.385 & 0.435 & 0.048 & 0.048 & 0.627 & 2.541 \\
Hermes AI & 0.321 & 0.301 & 0.369 & 0.433 & 0.048 & 0.095 & 0.348 & 1.474 \\
OpenClaw & 0.318 & 0.248 & 0.368 & 0.403 & 0.095 & 0.048 & 0.625 & 2.545 \\
EvoAgentX & 0.272 & 0.263 & 0.348 & 0.414 & 0.048 & 0.048 & 0.611 & 2.552 \\
\end{longtable}
}

Across both blocks, the central trustworthiness caveat is unchanged: agents can improve some aggregate metrics, but strict sustained-depeg detection remains weak. The binary TP/FN columns in Tables~\ref{tab:stress-recall} and~\ref{tab:stress-recall-507} make this limitation explicit.

\subsection{Anonymization and Leakage-Safety Audit}
\label{app:anonymization-leakage}

The anonymization ablation probes possible parametric-memory leakage by comparing predictions on the same numerical evidence with real stablecoin names and calendar dates versus anonymized coin aliases and relative time labels. This is an audit signal rather than a proof of no leakage. The matched 12-case design preserves the numerical prompt evidence while changing identifying metadata.

{\StableEvalCompactTable
\begin{longtable}{@{}p{0.18\linewidth}rrrr@{}}
\caption{Anonymization ablation summary on matched numerical evidence. Lower label-change and numeric-difference rates indicate less sensitivity to coin/date identifiers, but this audit does not prove absence of leakage.}
\label{tab:anonymization-ablation}\\
\toprule
\textbf{System} & \textbf{Cases} & \textbf{Label change} & \textbf{Mean $p$ diff.} & \textbf{Mean bps diff.} \\
\midrule
\endfirsthead
\toprule
\textbf{System} & \textbf{Cases} & \textbf{Label change} & \textbf{Mean $p$ diff.} & \textbf{Mean bps diff.} \\
\midrule
\endhead
\bottomrule
\endlastfoot
EvoAgentX & 12 & 0.1667 & 0.01667 & 7.917 \\
Hermes AI & 12 & 0.08333 & 0.02917 & 4.583 \\
LangGraph & 12 & 0.3333 & 0.06250 & 8.417 \\
Nanobot & 12 & 0.2500 & 0.04167 & 8.500 \\
NanoClaw & 12 & 0.2500 & 0.05000 & 14.170 \\
OpenClaw & 12 & 0.08333 & 0.02917 & 8.333 \\
\end{longtable}
}

\subsection{Calibration and Metric-Consistency Audits}
\label{app:calibration-consistency}

The calibration rule is frozen and applied after model output parsing. It does not change probabilities, predicted maximum deviations, cost, latency, token counts, hidden labels, or future data. The 507 release also includes saved metric-consistency and alignment audits under \texttt{results/stableeval\_507\_natural/metrics/verification/} and \texttt{results/stableeval\_507\_natural/reports/}. These files are used as audit artifacts rather than additional leaderboard tables, because the paper-facing values are already reported in Sections~\ref{app:stableeval-120} and~\ref{app:stableeval-507}.

\newpage
\section{Reproducibility Artifacts and Release Checklist}
\label{app:reproducibility}

This appendix is designed to function as an audit layer for the main paper. StableEval Arena preserves the intermediate artifacts needed to trace each reported result from case construction, to agent execution, to calibrated scoring, to final paper-facing metrics. The release package prioritizes frozen case lists, prompt packets, model and platform metadata, raw predictions, calibration rules, evaluator outputs, cost logs, and cross-variant summaries.

{\StableEvalArtifactTable
\begin{longtable}{@{}p{0.38\linewidth}p{0.54\linewidth}@{}}
\caption{Main reproducibility artifacts preserved for StableEval Arena. Paths are relative to the release root.}
\label{tab:reproducibility-artifacts}\\
\toprule
\textbf{Artifact} & \textbf{Purpose} \\
\midrule
\endfirsthead
\toprule
\textbf{Artifact} & \textbf{Purpose} \\
\midrule
\endhead
\bottomrule
\endlastfoot
\StableEvalPathCell{\texttt{data/benchmark/}\\\texttt{stableeval\_507\_natural/}\\\texttt{labels.csv}} & Full frozen 507-case label file used as the source label table for benchmark evaluation. \\
\StableEvalPathCell{\texttt{data/benchmark/}\\\texttt{stableeval\_507\_natural/}\\\texttt{case\_list.csv}} & Full 507-case natural-distribution case list. \\
\StableEvalPathCell{\texttt{data/benchmark/}\\\texttt{stableeval\_507\_natural/}\\\texttt{prompt\_packets.jsonl}} & Full 507-case risk-sensitive prompt-packet file containing lookback-only evidence and excluding hidden future labels. \\
\StableEvalPathCell{\texttt{data/benchmark/}\\\texttt{stableeval\_507\_natural/}\\\texttt{full\_hourly\_windows.csv.gz}} & Expanded frozen lookback and horizon rows used for audit and reconstruction. \\
\StableEvalPathCell{\texttt{data/benchmark/}\\\texttt{stableeval\_120\_enriched/}\\\texttt{prompt\_packets.jsonl}} & 120-case stress-enriched prompt packets. \\
\StableEvalPathCell{\texttt{results/stableeval\_120\_enriched/}\\\texttt{validation\_case\_list.csv}} & Selected 120-case stress-enriched validation set used for the main reported validation results. \\
\StableEvalPathCell{\texttt{prompts/templates/}\\\texttt{risk\_sensitive\_agent\_prompt.md}} & Frozen risk-sensitive agent prompt template shared across evaluated platforms. \\
\StableEvalPathCell{\texttt{results/stableeval\_120\_enriched/}\\\texttt{model\_backend.json}} & Frozen model-backend metadata for the 120-case validation protocol. \\
\StableEvalPathCell{\texttt{results/stableeval\_120\_enriched/}\\\texttt{calibration\_rule.md}} & Frozen calibrated-label mapping used to convert model risk estimates into stable/watch/stress labels. \\
\StableEvalPathCell{\texttt{results/stableeval\_120\_enriched/}\\\texttt{agent\_execution\_metadata.csv}} & Platform package, execution-mode, native-runtime, and adapter metadata. \\
\StableEvalPathCell{\texttt{results/stableeval\_120\_enriched/}\\\texttt{predictions\_base\_raw/}} & Saved first-pass 120-case raw agent predictions. \\
\StableEvalPathCell{\texttt{results/stableeval\_120\_enriched/}\\\texttt{predictions\_base\_calibrated/}} & Saved 120-case calibrated predictions used for the main validation table. \\
\StableEvalPathCell{\texttt{results/stableeval\_120\_enriched/}\\\texttt{evaluation\_base\_calibrated/}} & Evaluator outputs for the main 120-case validation result. \\
\StableEvalPathCell{\texttt{results/stableeval\_120\_enriched/}\\\texttt{final\_tables/}} & Paper-facing 120-case summary, stress, cost, and ablation tables. \\
\StableEvalPathCell{\texttt{results/stableeval\_120\_enriched/}\\\texttt{freeze\_manifest/}} & 120-case freeze manifest and checksum summary used to audit that later scaling work did not alter frozen validation artifacts. \\
\StableEvalPathCell{\texttt{results/stableeval\_507\_natural/}\\\texttt{raw\_predictions/}} & Saved raw 507-case agent predictions. \\
\StableEvalPathCell{\texttt{results/stableeval\_507\_natural/}\\\texttt{calibrated\_predictions/}} & Saved calibrated 507-case agent predictions. \\
\StableEvalPathCell{\texttt{results/stableeval\_507\_natural/}\\\texttt{baselines/}} & 507-case non-agent baseline predictions. \\
\StableEvalPathCell{\texttt{results/stableeval\_507\_natural/}\\\texttt{final\_tables/}\\\texttt{stableeval\_507\_natural\_v6\_}\\\texttt{leaderboard.csv}} & Canonical full 507-case leaderboard containing all six agentic systems and four non-agent baselines. \\
\StableEvalPathCell{\texttt{results/stableeval\_507\_natural/}\\\texttt{metrics/}} & Per-system 507-case metric outputs and with-actuals files used to reconstruct predictive, risk, reliability, cost, and binary depeg metrics. \\
\StableEvalPathCell{\texttt{results/stableeval\_507\_natural/}\\\texttt{costs/}} & 507-case cost, latency, and token accounting files for agentic systems and baselines. \\
\StableEvalPathCell{\texttt{results/stableeval\_507\_natural/}\\\texttt{metadata/calibration\_rule.md}} & Frozen post-hoc calibration rule applied to 507-case agent predictions. \\
\StableEvalPathCell{\texttt{results/stableeval\_507\_natural/}\\\texttt{metrics/verification/}} & 507-case case-label, calibration, metric-recomputation, and table-version consistency checks. \\
\StableEvalPathCell{\texttt{results/stableeval\_507\_natural/}\\\texttt{reports/}} & 120-versus-507 alignment and scaling-run interpretation reports. \\
\StableEvalPathCell{\texttt{protocols/}\\\texttt{stableeval\_507\_natural\_}\\\texttt{protocol.json}} & Frozen 507-case protocol descriptor, including backend, temperature, systems, prompt path, and canonical output paths. \\
\StableEvalPathCell{\texttt{scripts/evaluate/}\\\texttt{recompute\_stableeval\_120\_}\\\texttt{enriched.py}} & No-API recomputation script for the main 120-case saved-output metrics. \\
\StableEvalPathCell{\texttt{scripts/evaluate/}\\\texttt{recompute\_stableeval\_507\_}\\\texttt{natural.py}} & No-API recomputation script for the 507-case v6 leaderboard, stress audit, and cost/reliability summary. \\
\texttt{FILE\_MANIFEST.md} & Release-facing file manifest and checksums. \\
\texttt{REPRODUCIBILITY.md} & Reproduction guide separating saved-output checks from optional API-backed reruns. \\
\end{longtable}
}

\endgroup

\end{document}